\documentclass[10pt,twocolumn,twoside]{IEEEtran}
\usepackage{cite}
\usepackage[cmex10]{amsmath}
\usepackage{amssymb}
\usepackage{graphicx}
\usepackage{cases}
\usepackage{mdwmath}
\usepackage{mdwtab}
\usepackage{array}
\usepackage{url}
\usepackage{multirow}
\usepackage{booktabs}
\usepackage{threeparttable}
\usepackage{color}
\usepackage{amsmath}
\usepackage{changepage}

\def\ie{\emph{i.e.}}
\def\eg{\emph{e.g.}}
\def\etal{{\em et al.}}

\newcommand{\bl}[1]{\textbf{#1}}
\newcommand{\mc}[1]{\mathcal{#1}}

\graphicspath{{./figs/}}
\usepackage{color}

\allowdisplaybreaks

\begin{document}
%
\title{Salient Object Detection with Purificatory Mechanism and Structural Similarity Loss}
%
%
%

\author{Jia Li,~\IEEEmembership{Senior Member,~IEEE,}
        Jinming Su,
        Changqun Xia,
        Mingcan Ma,
        and~Yonghong Tian,~\IEEEmembership{Senior Member,~IEEE}
\thanks{
J. Li, M. Ma and J. Su are with the State Key Laboratory of Virtual Reality Technology and Systems, School of Computer Science and Engineering, Beihang University, Beijing, 100191, China.

Y. Tian is with the Department of Computer Science and Technology, Peking University, Beijing,100871, China.

J. Li, C. Xia, M. Ma and Y. Tian are with Peng Cheng Laboratory, Shenzhen, 518000, China.}

\thanks{Correspondence should be addressed to Changqun Xia and Jia Li. E-mail: xiachq@pcl.ac.cn, jiali@buaa.edu.cn. Website: http://cvteam.net }
}

%
%

\markboth{IEEE Transactions on Image Processing}
{IEEE Transactions on Image Processing}

%



\maketitle

\begin{abstract}
Image-based salient object detection has made great progress over the past decades, especially after the revival of deep neural networks. By the aid of attention mechanisms to weight the image features adaptively, recent advanced deep learning-based models encourage the predicted results to approximate the ground-truth masks with as large predictable areas as possible, thus achieving the state-of-the-art performance. However, these methods do not pay enough attention to small areas prone to misprediction. In this way, it is still tough to accurately locate salient objects due to the existence of regions with indistinguishable foreground and background and regions with complex or fine structures. To address these problems, we propose a novel convolutional neural network with purificatory mechanism and structural similarity loss. Specifically, in order to better locate preliminary salient objects, we first introduce the promotion attention, which is based on spatial and channel attention mechanisms to promote attention to salient regions. Subsequently, for the purpose of restoring the indistinguishable regions that can be regarded as error-prone regions of one model, we propose the rectification attention, which is learned from the areas of wrong prediction and guide the network to focus on error-prone regions thus rectifying errors. Through these two attentions, we use the \emph{Purificatory Mechanism} to impose strict weights with different regions of the whole salient objects and purify results from hard-to-distinguish regions, thus accurately predicting the locations and details of salient objects. In addition to paying different attention to these hard-to-distinguish regions, we also consider the structural constraints on complex regions and propose the \emph{Structural Similarity Loss}. The proposed loss models the region-level pair-wise relationship between regions to assist these regions to calibrate their own saliency values. In experiments, the proposed purificatory mechanism and structural similarity loss can both effectively improve the performance, and the proposed approach outperforms 19 state-of-the-art methods on six datasets with a notable margin. Also, the proposed method is efficient and runs at over 27FPS on a single NVIDIA 1080Ti GPU.
\end{abstract}

%
\IEEEpeerreviewmaketitle

\section{Introduction}
\IEEEPARstart{V}{isual} saliency plays an essential role in the human vision system, which guides human beings to look at the most important information from visual scenes and can be well referred to as the allocation of cognitive resources on information~\cite{james1890principles,li2014visual}. To model this mechanism of visual saliency, there are two main research branches in computer vision: fixation prediction~\cite{itti1998model} and salient object detection~\cite{borji2015salient}. This work focuses on the second one (\ie, salient object detection, abbreviated as SOD), which aims to detect and segment the most visually distinctive objects.
Over the past years, SOD has made significant progress, and it is also used as an important preliminary step for various vision tasks, such as object recognition~\cite{ren2014region}, tracking~\cite{hong2015online} and image parsing~\cite{lai2016saliency}.

\begin{figure}[t]
\centering
\includegraphics[width=1\columnwidth]{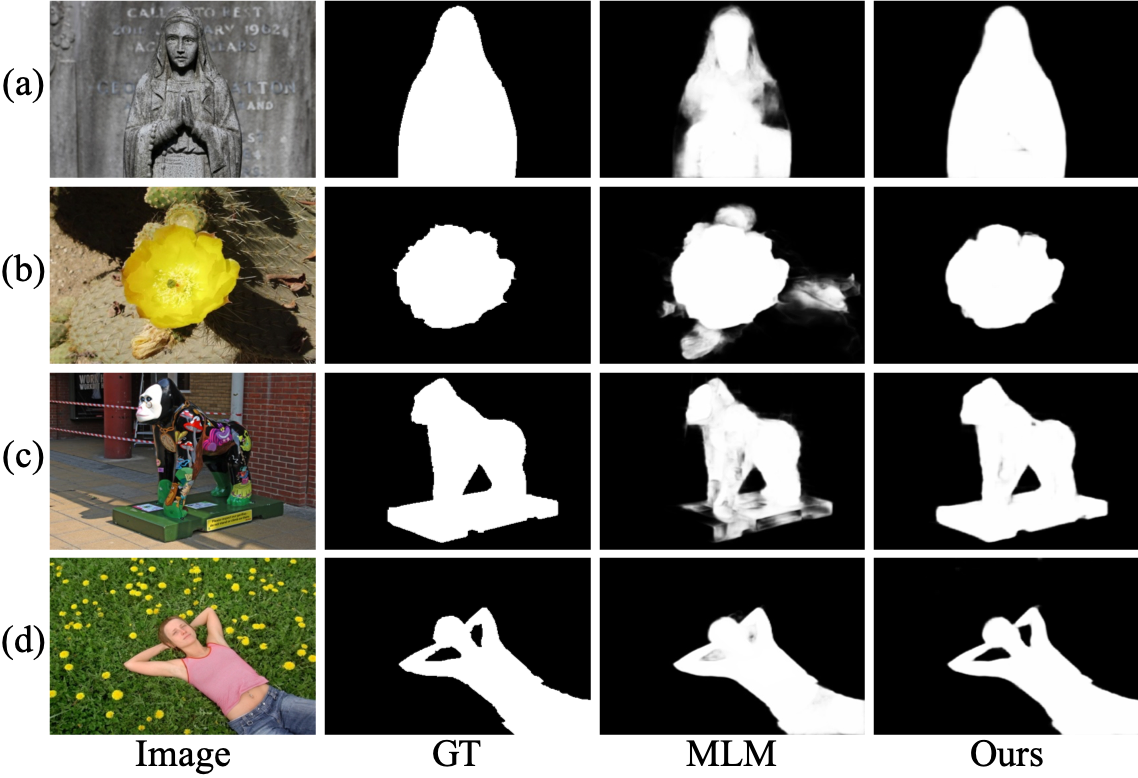}
\caption{Difficulties that hinder the development of SOD. In (a)(b), these usually exist regions with similar foreground and background, which confuses the models to cause wrong predictions. In (c)(d),  complex structures caused by complex illumination or color and fine hollows make it difficult to maintain the structural integrity and clarity. Images and ground-truth masks (GT) are from ECSSD~\cite{yan2013hierarchical}. Results are generated by MLM~\cite{wu2019mutual}  and our approach.}
\label{fig:motivation}
\end{figure}

To address the SOD task, lots of learning-based methods~\cite{lee2016deep,liu2016dhsnet,hou2017deeply,luo2017non,zhang2017amulet,wang2017stagewise,zhang2017learning,chen2017look,zhang2018bi,zeng2018learning,wang2018salient,wang2019iterative,su2019selectivity} have been proposed in recent years, achieving impressive performance on existing benchmark datasets~\cite{yan2013hierarchical, yang2013saliency, li2014secrets, li2015visual, wang2017learning, xia2017and}.
However, there still exist two difficulties that hinder the development of SOD. First, it is hard to distinguish these regions with similar foreground and background. As shown in Fig.~\ref{fig:motivation}(a)(b), these regions usually confuse the models to cause wrong predictions, and we named these regions as ``error-prone region'' of models. Second, it is difficult to restore the complex or fine structures. As displayed in Fig.~\ref{fig:motivation}(c)(d), the complex structures (\eg, caused by complex illumination and color) and fine structures (\eg,  hollows) make it difficult to maintain the structural integrity and clarity. These two problems are especially difficult to deal with for existing SOD methods and greatly hinder the performance of SOD. Due to these difficulties, SOD remains a challenging vision task.

To deal with the first difficulty, some methods~\cite{liu2018picanet,chen2018reverse,zhang2019capsal,wu2019cascaded,zhao2019pyramid,wangwenguan2019salient} adopt attention mechanisms to weight the features adaptively to focus on salient regions. For examples, Zhang~\etal~\cite{zhang2018progressive} introduced an attention guided network to integrate multi-level contextual information by utilizing global and local attentions, consistently improving saliency detection performance. Chen~\etal~\cite{chen2018reverse} proposed the reverse attention to guide the side-output residual learning in a top-down manner to restore the salient object parts and details.
For these methods, although different forms of features are effectively aggregated, the overall goal is to make the prediction results approach ground-truth masks with as larger an intersection as possible, which improves the accuracy of the area that is easy to predict. However, these methods mainly focuses on improving the correctness of large predictable areas, but don't pay enough attention to small error-prone areas.
To address the second problem, methods~\cite{wang2018detect,li2018contour,wu2019mutual,liu2019simple,feng2019attentive,qin2019basnet,wangwenguan2019salient,su2019selectivity} consider to solve the problem of inaccurate boundaries. For example, Wang~\etal~\cite{wang2018detect} proposed a local boundary refinement network to recover object boundaries by learning the local contextual information for each spatial position. Wu~\etal~\cite{wu2019mutual} also adopted the foreground contour and edge to guide each other, thereby leading to precise foreground contour prediction and reducing the local noises. In these methods, some special boundary branches and losses are proposed to attend boundaries or local details. In this way, these methods mainly take account of the unary supervision to deal with the complex and fine structures. But for many complex nad fine structures that are influenced by the context, it is difficult to accurately restore only considering the unary information, which only considers the correlation at the pixel level but not at the regional level.


Inspired by these observations and analyses, we propose a novel convolutional neural network with purificatory mechanism and structural similarity loss for image-based SOD. In the network, we propose the  purificatory mechanism to purify salient objects by promoting predictable regions and rectifying indistinguishable regions. In this mechanism, we first introduce a simple but effective promotion attention based on spatial and channel attention mechanisms to provide the promotion ability, which assists to locate preliminary salient objects. Next, we propose a novel rectification attention, which predicts the error-prone areas and guides the network to pay more attention to these areas to rectify errors from the aspect of features and losses. These two attentions are used to impose strict weights with different regions of the whole salient objects and formed the purificatory mechanism. In addition, in order to better restore the complex or fine structures of salient objects, we propose a novel structural similarity loss to model and constrain the structural relation on complex regions for better calibrating the saliency values of regions, which can be regarded as an effective supplement to the pixel-level unary constraint. The purificatory mechanism and structural similarity loss are integrated in a progressive manner to pop-out salient objects. Experimental results on six public benchmark datasets verify the effectiveness of our method which consistently outperforms 19 state-of-the-art SOD models with a notable margin. Moreover, the proposed method is efficient and runs at about 27FPS on a single NVIDIA 1080Ti GPU.

The main contributions of this paper include:
\begin{enumerate}
\item we propose a novel \emph{Purificatory Mechanism}, which purifies salient objects by promoting predictable regions and rectifying indistinguishable regions;
\item we introduce a novel \emph{Structural Similarity Loss} to restore the complex or fine structures of salient objects, which constrains region-level pair-wise relationship between regions to be as a supplement to the pixel-level unary constraints, assisting regions to calibrate their own saliency values;
\item we conduct comprehensive experiments and the results verify the effectiveness of our proposed method which consistently outperforms 19 state-of-the-art algorithms on six datasets with a fast prediction.
\end{enumerate}

The rest of this paper is organized as follows: Section II reviews the recent development of salient object detection, attention-based SOD methods and boundary-aware SOD methods. Section III presents the  purificatory network in detail. Section IV presents the proposed structural similarity loss. In Section V, we evaluate the proposed model, and compare it with the state-of-the-art methods to validate the effectiveness of the model. We conclude the paper in Section VI.

\section{Related Work}
In this section, we review the related works in three aspects. At the beginning, some representative salient object detection methods are introduced. Next, we present attention mechanisms and attention-based SOD methods. Next, we review the boundary-aware SOD methods.

\subsection{Salient Object Detection}
Hundreds of image-based SOD methods have been proposed in the past decades. Early methods mainly adopted hand-crafted local and global visual features as well as heuristic saliency priors such as color difference \cite{achanta2009frequency}, distance transformation \cite{tu2016real} and local/global contrast \cite{klein2011center,cheng2015global}. More details about the traditional methods can be found in the survey \cite{borji2015salient}.

With the development of deep learning, many deep neural networks (DNNs) based methods~\cite{lee2016deep,liu2016dhsnet,hou2017deeply,luo2017non,zhang2017amulet,wang2017stagewise,zhang2017learning,chen2017look,zhang2018bi,zeng2018learning,wang2018salient,wang2019iterative,su2019selectivity} have been proposed for SOD. Lots of deep models are devoted to fully utilizing the feature integration to enhance the performance of DNNs. For example, Lee~\etal~\cite{lee2016deep} proposed to compare the low-level features with other parts of an image to form a low-level distance map. Then they concatenated the encoded low-level distance map and high-level features extracted by VGG~\cite{simonyan2015very}, and connect them to a DNN-based classifier to evaluation the saliency of a query region. Liu~\etal~\cite{liu2016dhsnet} presented a DHSNet that first made a coarse global prediction by learning various global structured saliency cures and then adopted a recurrent convolutional neural network to refine the details of saliency maps by integrating local contexts step by step, which worked in a global to local and coarse to fine manner.

In addition, Hou~\etal~\cite{hou2017deeply} introduced short connections to the skip-layer structures, which provided rich multi-scale feature maps at each layer, performing salient object detection. Luo~\etal~\cite{luo2017non} proposed a convolutional neural network by combining global and local information through a multi-resolution $4 \times 5$ grid structure to simplify the model architecture and speed up the computation. Zhang~\etal~\cite{zhang2017amulet} adopted a framework to aggregate multi-level convolutional features into multiple resolutions, which were then combined to predict saliency maps in a recursive manner. Wang~\etal~\cite{wang2017stagewise} proposed a pyramid pooling module and a multi-stage refinement mechanism to gather contextual information and stage-wise results, respectively. Zhang~\etal~\cite{zhang2017learning} utilized the deep uncertain convolutional features and proposed a reformulated dropout after specific convolutional layers to construct an uncertain ensemble of internal feature units. Chen~\etal~\cite{chen2017look} incorporated human fixation with semantic information to simulate the human annotation process to form two-stream fixation-semantic CNNs, which were fused by an inception-segmentation module. Zhang~\etal~\cite{zhang2018bi} proposed a novel bi-directional message passing model to integrate multi-level features for SOD.

These methods usually integrate multi-scale and multi-level feature by complex structures to improve the representation ability of DNNs. To simply and effectively integrate these features, we add lateral connections to transfer encoded features to assist the decoder and adopt a top-down architecture to propagate high-level semantics to low-level details as guide of locating salient objects as well as restoring object details.

\subsection{Attention-based Methods}
Attention mechanism of DNNs is inspired from human perception process, which weights the features to encourage one model to focus on important information. The mechanism was first applied in machine translation~\cite{bahdanau2015neural} and then widely used in the field of computer vision due to its effectiveness. For example, Mnih~\etal~\cite{mnih2014recurrent} applied an attention-based model to image classification tasks. In~\cite{chen2017sca}, SCA-CNN that incorporated spatial and channel-wise attention mechanisms in a CNN are proposed to modulate the sentence generation context in multi-layer feature maps, encoding where and what the visual attention is, for the task of image captioning. Chu~\etal~\cite{chu2017multi} combined the holistic attention model focusing on the global consistency and the body part attention model focusing on detailed descriptions for human pose estimation. Fu~\etal~\cite{fu2019dual} proposed the dual attention network that adopted the position attention module aggregated the feat at each position and the channel attention module emphasizes interdependent channel maps for scene segmentation. Woo~\etal~\cite{woo2018cbam} proposed Convolutional Block Attention Module (CBAM) to efficiently help the information flow within the network by learning which information to modality or suppress.

Due to the effectiveness of attention mechanisms for feature enhancement, they are also applied to saliency detection. Liu~\etal~\cite{liu2018picanet} proposed a pixel-wise contextual attention network to learn to attend to informative context locations for each pixel by two attentions: global attention and local attention, guiding the network learning to attend global and local contexts, respectively.  Feng~\etal~\cite{feng2019attentive} designed the attentive feedback modules to control the message passing between encoder and decoder blocks, which was considered an opportunity for error corrections. Zhang~\etal~\cite{zhang2019capsal} leveraged captioning to boost semantics for salient object detection and introduced a textual attention mechanism to weight the importance of each word in the caption. In~\cite{wu2019cascaded}, a holistic attention module was proposed to enlarge the coverage area of these initial saliency maps since some objects in complex scenes were hard to be completely segmented. Zhao and Wu~\cite{zhao2019pyramid} presented a pyramid feature attention network to enhance the high-level context features and the low-level spatial structural features. Wang~\etal~\cite{wangwenguan2019salient} proposed a pyramid attention structure to offer the representation ability of the corresponding network layer with an enlarged receptive field.

In the above methods, attention mechanisms (spatial attention and channel attention) are used to enhance the localization and awareness of salient objects. These attentions play good roles in promoting feature attention to salient regions, but lacks attention to small regions prone to mis-prediction. Unlike these methods, we propose the purificatory mechanism, which introduce two novel attentions: promotion attention and rectification attention. The first attention is dedicated to promoting the feature representation of salient regions, while the second one is dedicated to rectifying the features of error-prone regions.

\subsection{Boundary-aware Methods}
Some methods~\cite{wang2018detect,li2018contour,wu2019mutual,liu2019simple,feng2019attentive,qin2019basnet,wangwenguan2019salient,su2019selectivity} consider that unclear object boundaries and inconsistent local details are important factors affecting performance of SOD.  Li~\etal~\cite{li2018contour} considered contours as useful priors and proposed to facilitate feature learning in SOD by transferring knowledge from an existing contour detection model. In~\cite{liu2019simple}, an edge detection branch was used to assist the deep neural network to further sharpen the details of salient objects by joint training. Feng~\etal~\cite{feng2019attentive} presented a boundary-enhanced loss for learning fine boundaries and worked with the cross-entropy loss for saliency detection. Qin~\etal~\cite{qin2019basnet} also proposed a loss for boundary-aware SOD and the loss guided the network to learn in three levels: pixel-level, patch-level and map-level. Besides, more effective loss functions, such as meas intersection-over--union loss. weighted binary cross-entropy loss and affinity field matching loss, have been made to capture the quality factors for salient object detection tasks~\cite{wang2019salientera}. In~\cite{wangwenguan2019salient}, a salient edge detection module is introduced to emphasize on the importance of salient edge information, encouraging better edge-preserving SOD. And Su~\etal~\cite{su2019selectivity} proposed a boundary-aware network, which split salient objects into boundaries and interiors, extracted features from different regions to ensure the representation of each region, and then fused to obtain good results.

\begin{figure*}[t]
\centering
\includegraphics[width=1\textwidth]{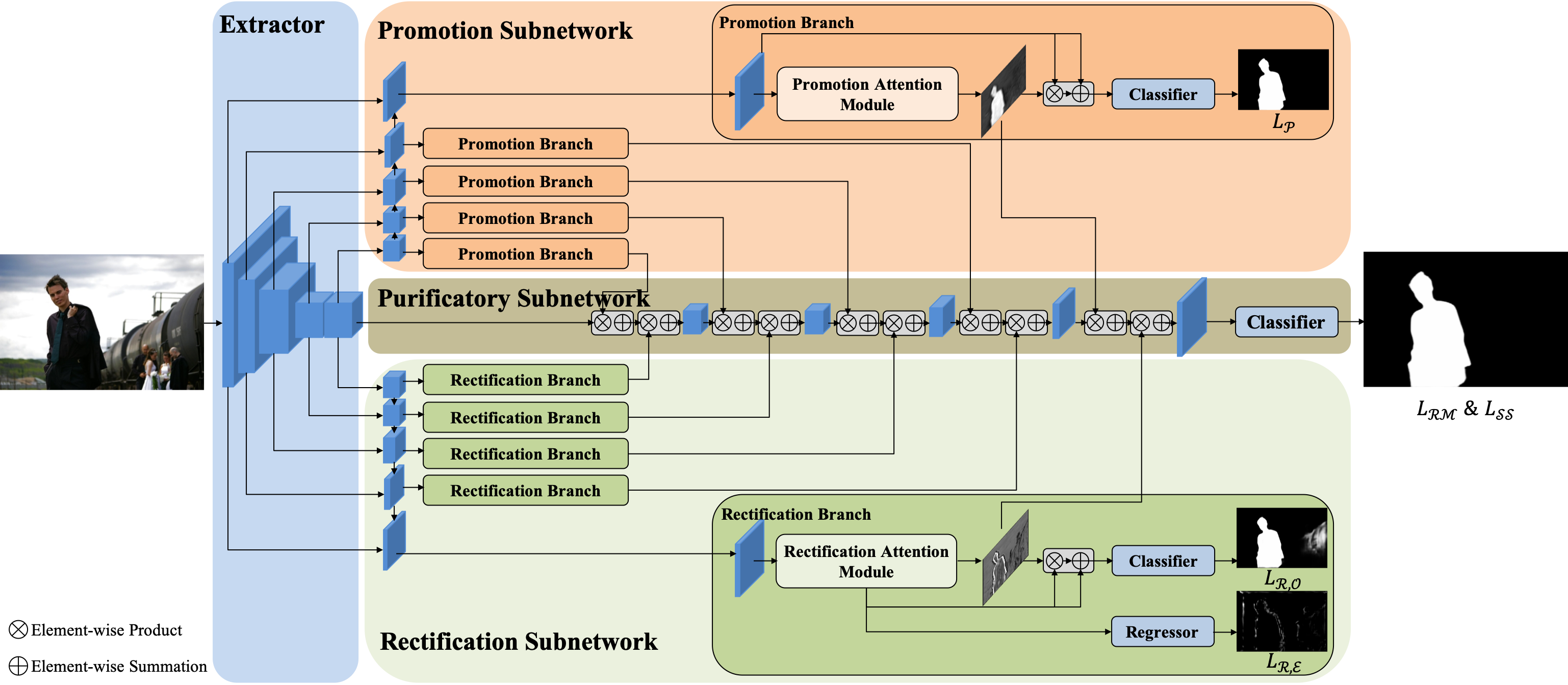}
\caption{The framework of our approach. We first extract the common features by extractor, which provides the features for the other three subnetworks. In detail, the promotion subnetwork produces promotion attention to guide the model to focus on salient regions, and the rectification subnetwork give the rectification attention for rectifying the errors. These two kind of attentions are combined to formed the purificatory mechanism, which is integrated in the purificatory subnetwork to refine the prediction of salient objects progressively.}
\label{fig:framework}
\end{figure*}

These methods usually utilize some special boundary branch and loss to attend boundaries or local details. But for many complex and fine structures that are influenced by the context, it is difficult to accurately restore only considering the unary information. Our method differs with these methods by introducing the structural similarity loss, which models and constrains the pair-wise structural relation on complex regions for better calibrating the saliency values of regions and is an effective supplement to the pixel-level unary constraint.

\section{Purificatory Network}
To address these problems (\ie, indistinguishable regions and complex structures), we propose a novel purificatory network (denoted as \textbf{PurNet}) for SOD. In this method, different regions are attended by corresponding attentions, \ie, promotion attention and rectification attention. The first one is to promote attention in salient regions and the second one aims to rectify errors for salient regions. In terms of the architecture, the network includes four parts: the feature extractor, the promotion subnetwork, the rectification subnetwork and the purificatory subnetwork. In this section, we first overview the whole purificatory network and then introduce each part separately. Details of the proposed approach are described as follows.

\subsection{Overview}
A diagram of the top-down architecture with feature transferring and utilization is as shown in Fig.~\ref{fig:backbone}, the proposed PurNet has a top-down basic architecture with lateral connections, which is used by the feature pyramid network (FPN)~\cite{lin2017feature} based on the encoder-decoder form. In our method, PurNet consists of four parts, and the first part (\ie, the extractor) provides the common features for the other three ones (regarded as decoders).
Each of the rest three parts forms an encoder-decoder relation with the feature extractor, and decodes the received features respectively. In the rest three decoders, the promotion subnetwork is used to provide the promotion features which is utilized to improve the localization ability and semantic information for salient regions and the rectification subnetwork provides rectification features which  can provide the rectification attention for rectifying the errors, while the purificatory subnetwork uses the purificatory mechanism to refine the prediction of SOD progressively.

\subsection{Feature Extractor}
\begin{figure}[t]
\centering
\includegraphics[width=1\columnwidth]{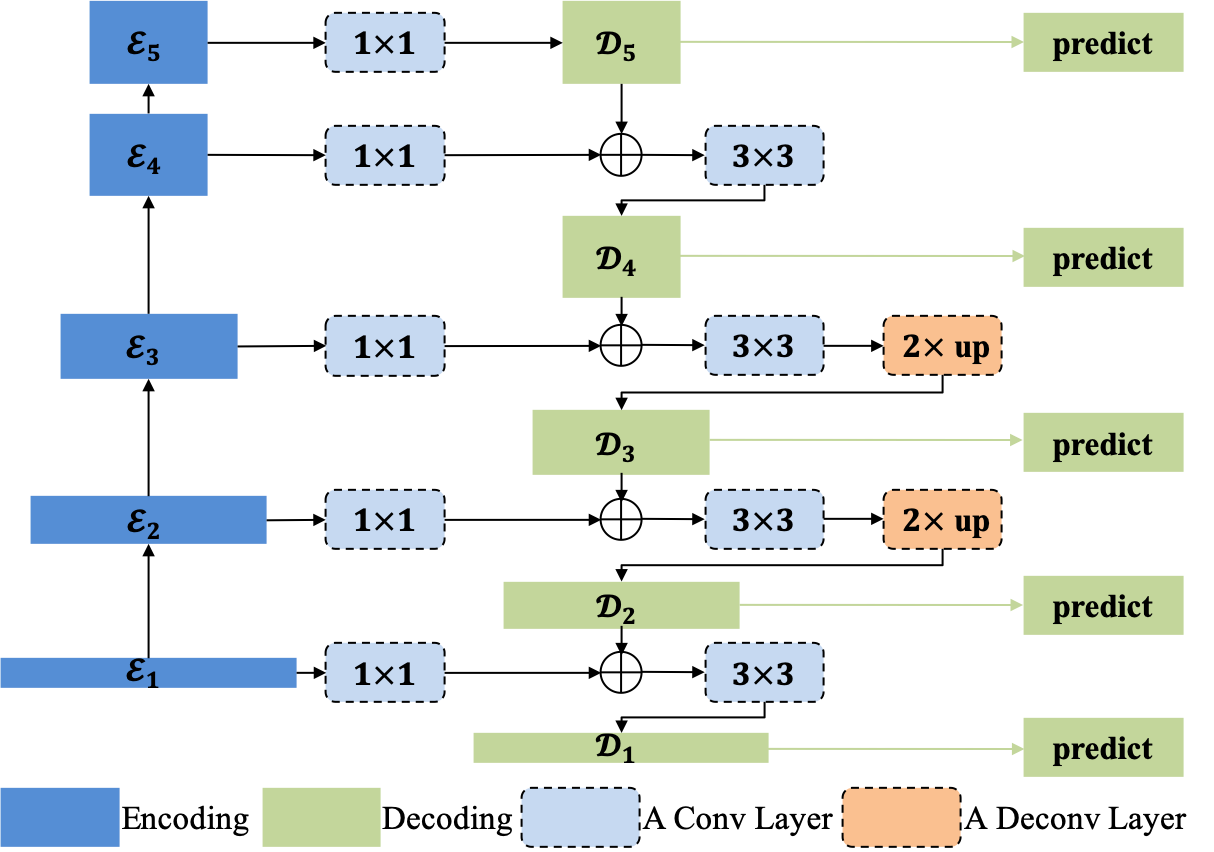}
\caption{The backbone of feature extractor. We adopt five residual modules for encoding, lateral connections to transfer the features of each encoding module to the decoder by a convolution layer with 128 kernels of $1 \times 1$ for utilizing multi-level and multi-scale features, and convolution layers with 128 kernels of $3 \times 3$ followed by a $2 \times$ upsampling deconvolution layer for decoding and restoring features.}
\label{fig:backbone}
\end{figure}

To see the Fig.~\ref{fig:backbone}, the purificatory network tasks ResNet-50~\cite{he2016deep} as the feature extractor, which is modified to remove the last global pooling and fully connected layers for the pixel-level prediction task. Feature extractor has five residual modules for encoding, named as $\mc{E}_{1}(\pi_1), \dots, \mc{E}_{5}(\pi_5)$ with parameters $\pi_1, \dots, \pi_5$. To obtain larger feature maps, the strides of all convolutional layers belonging to last residual modules $\mc{E}_{5}$ are set to 1. To further enlarge the receptive fields of high-level features, we set the dilation rates~\cite{yu2015multi} to 2 and 4 for convolution layers in $\mc{E}_{4}$ and $\mc{E}_{5}$, respectively. For a $H \times W$ input images, a $\frac{H}{16} \times \frac{W}{16}$ feature map is output by the feature extractor.

In order to integrate multi-level and multi-scale features, we adopt lateral connections to transfer the features of each encoding module to the decoder by a convolution layer with 128 kernels of $1 \times 1$, which also compresses the channels of high-level features for later processing and integration. In addition, we use a top-down architecture to propagate high-level semantics to low-level details as guide of locating salient objects as well as restoring object details. In this architecture, features from same-level encoding feature and higher-level decoding features are added, and a convolution layer with 128 kernels of $3 \times 3$ is used to decode these features. We use learnable deconvolution to perform $2 \times$ upsampling to align and restore features.

For the following three subnetworks (\ie, the promotion, rectification and purificatory subnetworks), there is a set of learned decoding features $\mathcal{D}_i, i \in \{1,\dots,5\}$, respectively. The three subnetworks mainly process these decoding features and predict the corresponding expected results.

\subsection{Promotion Subnetwork}

\subsubsection{Promotion Attention}
In general, when there are some distractions in the background, the location of salient objects is difficult to be detected as shown in Fig.~\ref{fig:motivation}(a)(b). Some methods~\cite{liu2018picanet,zhang2018progressive,chen2018reverse} consider to make their models focus on the salient regions by spatial attention and channel attention mechanisms. In these two mechanisms, the former can be used to enhance the localization capability, and the latter aims to enhance semantic information~\cite{zhao2019pyramid}.
For example,CBAM~\cite{woo2018cbam} and PAGRN~\cite{zhang2018progressive} adopt the cascade way to reconcile spatial and channel information and have has proven to be effective.
However, this way emphasizes the sequence of spatial and channel information in transmission, which will cause the loss of information in some complex scenes. In order to capture the contextual information in spatial and channel dimension, we pay more attention to the balance and reinforce of independent spatial information and channel information. Therefore, we propose a simple but effective parallel structure to provide the promotion ability.

\begin{figure}[t]
\centering
\includegraphics[width=1\columnwidth]{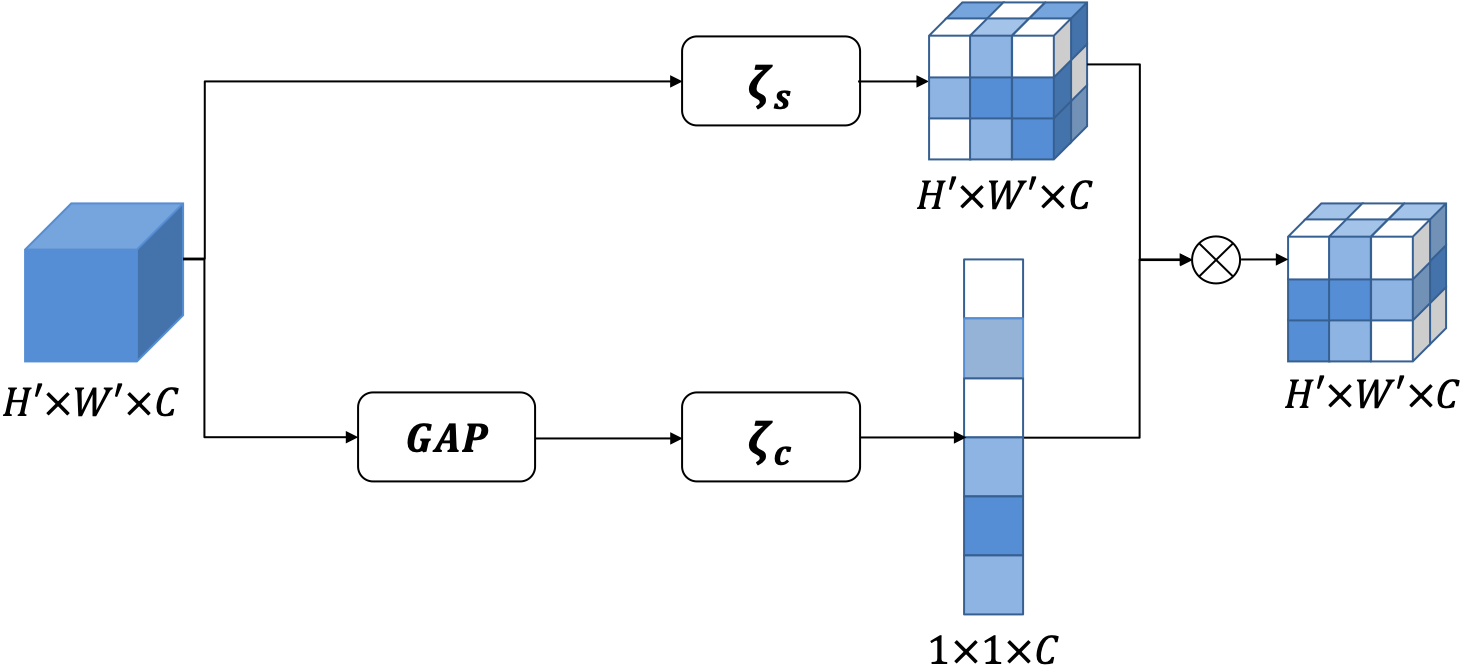}
\caption{The structure of the promotion attention module. The Softmax operation on spatial dimension ($\zeta_{s}$) is used to extract the spatial attention, and global average pooling ($GAP$) followed by the channel Softmax operation ($\zeta_{c}$) is used to obtain channel attention. The two attentions are multiplied as the promotion attention.}
\label{fig:promotion_attention}
\end{figure}

We present the structure of the promotion attention module as depicted in Fig.~\ref{fig:promotion_attention}. This moduel is based on existing spatial and channel attention without additional parameters. We denote input convolutional features as $\text{F}_{\mathcal{P}} \in \mathbb{R}^{H' \times W' \times C}$. The promotion attention is generated as follows:
\begin{equation}\label{eq:promotion_attention}
\text{A}_{{\mathcal{P}}} = \zeta_{s}(\text{F}_{{\mathcal{P}}}) \otimes \zeta_{c} (GAP(\text{F}_{{\mathcal{P}}})),
\end{equation}
where $\zeta_{s}(\cdot)$ and $\zeta_{c}(\cdot)$ denotes the Softmax operation on the spatial and channel dimension respectively, $GAP(\cdot)$ is the operation of global average pooling, and $\otimes$ represents element-wise product.

\begin{figure}[t]
\centering
\includegraphics[width=1\columnwidth]{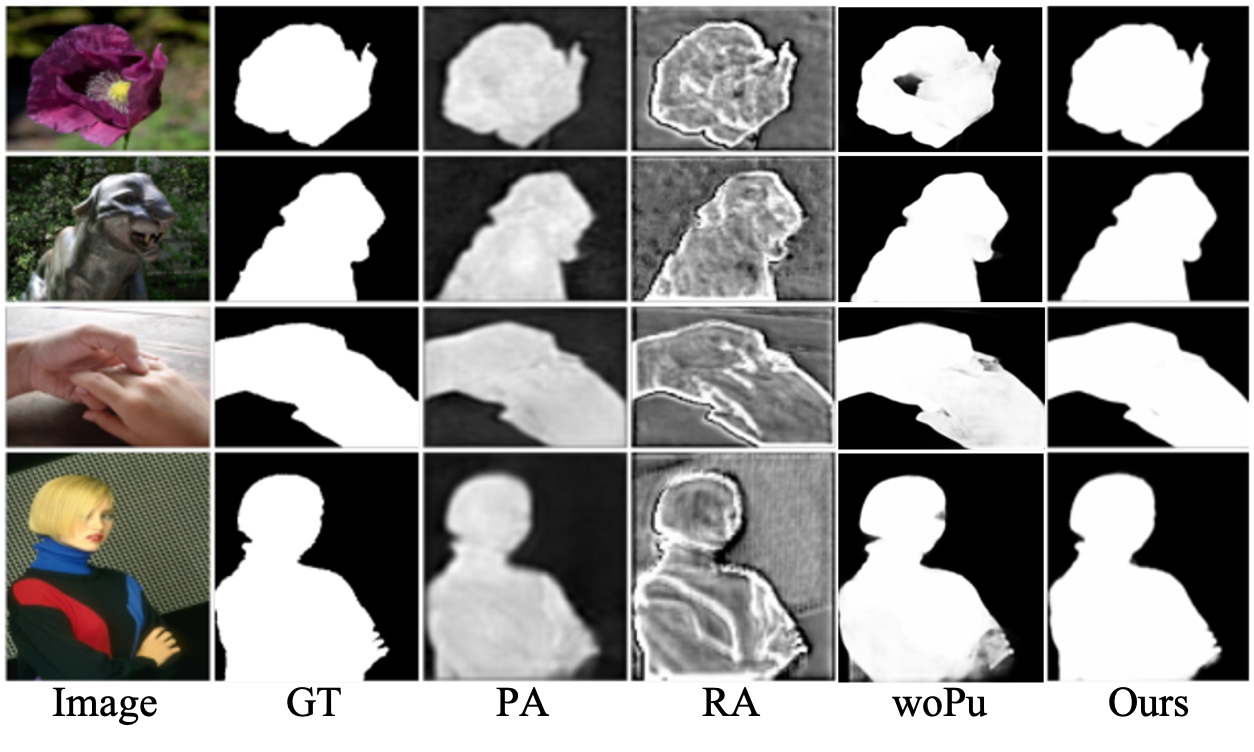}
\caption{Visual examples of the purificatory mechanism. GT: ground-truth mask, PA: the promotion attention, RA: the rectification attention, woPu: the prediction without purificatory mechanism, Ours: prediction of our approach.}
\label{fig:PM_results}
\end{figure}

In Eq.~(\ref{eq:promotion_attention}), the first item $\zeta_{s}(\text{F}_{{\mathcal{P}}}) $ is spatial attention, where a Softmax operation on spatial dimension  is directly conducted to obtain the spatial weights, and the second item $\zeta_{c} (GAP(\text{F}_{{\mathcal{P}}}))$ is the channel attention, where global average pooling is adopted to remove the effect of spatial for getting a vector of length $C$ followed by a Softmax operation on channel dimension to obtain the channel weights. Then, the attentions of spatial and channel dimension decouple and they are integrated by an element-wise product operation. In this manner, the advantage of our parallel structure lies in the adaptive allocation of spatial and channel information weights, thus avoiding artificial design and interference of different information weights and leading to locate preliminary salient objects more efficiently. Some visual examples can be found in the third column of Fig.~\ref{fig:PM_results}.

\subsubsection{Subnetwork}
As shown in Fig.~\ref{fig:framework}, the promotion attention module exists in the promotion subnetwork. In the promotion subnetwork, features from the five lateral connections of the feature extractor are firstly decoded. And then, each branch processes one of different level decoding features. For each branch, we represent input decoding convolutional features as $\text{F}_{\mathcal{P}} \in \mathbb{R}^{H \times W \times C}$ (the same features $\text{F}_{\mathcal{P}}$ in Eq.~(\ref{eq:promotion_attention})). Then the promotion attention module is utilized to weight the input features $\text{F}_\mathcal{P}$ by the following operation:
\begin{equation}
\text{M}_{\mathcal{P}} = \text{F}_{\mathcal{P}} \otimes \text{A}_{\mathcal{P}} + \text{F}_{\mathcal{P}}.
\end{equation}
The generated features $\text{M}_\mathcal{P}$ is then classified by a classifier, which includes two convolution layers with 128 kernels of $3 \times 3$ and $1 \times 1$, and one kernel of $1 \times 1$ followed by a Sigmoid and upsampling operation.

For the sake of simplification, these five branches of the promotion subnetwork are denoted as $\phi^{(i)}_{\mathcal{P}}(\pi^{(i)}_{\mathcal{P}}) \in [0, 1]^{H \times W \times 1}, i \in \{1, \dots, 5\}$, where $\pi^{(i)}_{\mathcal{P}}$ is the set of parameters of $\phi^{(i)}_{\mathcal{P}}$. As mentioned earlier, the promotion subnetwork aims to learn the promotion attention. To achieve this, we expect the output of the promotion attention module to approximate the ground-truth masks of SOD (represented as $G$) by minimizing the loss:
\begin{equation}
L_{\mathcal{P}} = \sum^5_{i=1} BCE(\phi^{(i)}_{\mathcal{P}}(\pi^{(i)}_{\mathcal{P}}), G),
\label{eq:promotion_network_loss}
\end{equation}
where $BCE(\cdot, \cdot)$ means the binary cross-entropy loss function with the following formulation:
\begin{equation} \label{eq:binary_cross_entropy}
BCE(P, G) = -\sum^{H \times W}_{i}(G_{i}\mathrm{log}P_i + (1 - G_{i})\mathrm{log}(1 - P_i)),
\end{equation}
where $P_{i}$ and $G_i$ represents the $i$th pixel of predicted maps and ground-truth masks of salient objects, respectively.

By taking multi-level lateral features from feature extractor as input, the promotion subnetwork can learning the promotion attention in multi-scale manner, which is fed to the purificatory subnetwork to promote attention to salient regions and demonstrates the strong promotion ability for SOD.

\subsection{Rectification Subnetwork}
In order to restore the structure of confusing or complicated areas (these areas can be regarded as error-prone regions of one model), we present the rectification mechanism, which is obtained by predicting the error-prone regions of the model. We are paying more attention to these areas and at the same time imposing stricter constraints, thus rectifying these errors.
\subsubsection{Rectification Attention}
As shown in Fig.~\ref{fig:motivation}(a)(b), it is difficult to accurately define the attributes and locations of some error-prone areas (\eg, salient regions confused with background). Therefore, we propose the rectification attention to guide the model to focus on these error-prone areas for error correction.

The structure of rectification attention module is shown in Fig.~\ref{fig:rectification_attention}. This module exists in the rectification branch. We represent the input features as $\text{F}_{\mathcal{R}} \in \mathbb{R}^{H \times W \times C}$. Then two parallel convolution branches are used to process the input features, where each branch has two convolution layers with 128 kernels of $3 \times 3$ followed by a convolution layer with one kernel of $1 \times 1$. We denote the outputs of these two branches as $\text{F}_{\mathcal{R, G}}$ and $\text{F}_{\mathcal{R, O}}$, which mean features of gross regions and object regions (named as gross branches and object branches). The gross features represent potential comprehensive features, while object feature represents predictable features in the object body, and their difference represents mispredicted features. Therefore, we use the subtraction ($\text{F}_{\mathcal{R, E}}$) of  $\text{F}_{\mathcal{R, G}}$ and $\text{F}_{\mathcal{R, O}}$ to be as the features of error-prone regions. Next, the rectification attention is generated as follows:
\begin{equation}
\text{A}_{\mathcal{R}} = \tau(\text{F}_{\mathcal{R, E}}),
\end{equation}
where $\text{F}_{\mathcal{R, E}} = \text{F}_{\mathcal{R, G}} - \text{F}_{\mathcal{R, O}}$ and $\tau(\cdot)$ is the Tanh function, which maps the features into range of $[-1, 1]$ to obtain the rectification attention. The rectification provides the attention to error-prone regions, which are important but almost undiscovered information for SOD. Some examples of rectification attention are shown in the forth column of Fig.~\ref{fig:PM_results}.

\subsubsection{Subnetwork}
\begin{figure}[t]
\centering
\includegraphics[width=1\columnwidth]{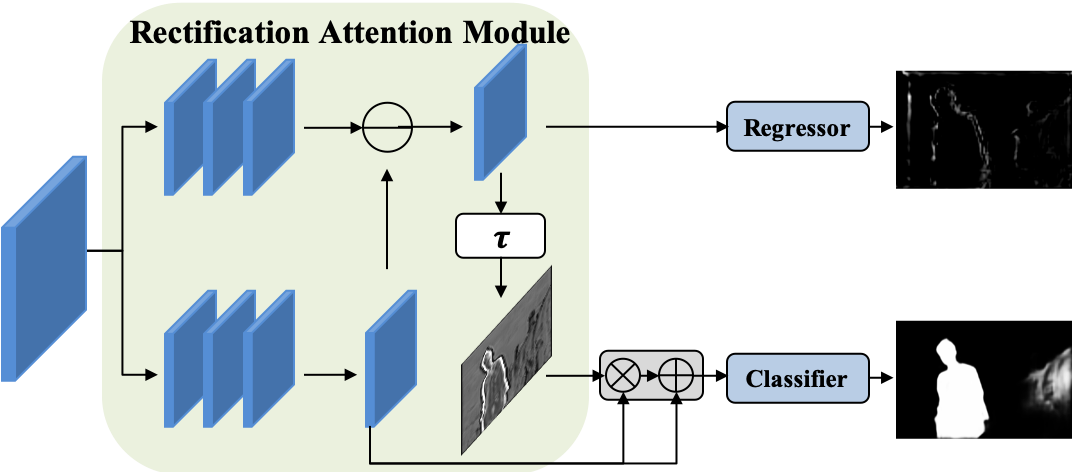}
\caption{The structure of the rectification branch. $\tau(\cdot)$ is the Tanh function. The output of the Classifier predicts salient objects and the one of the Regressor predicts errors in the saliency prediction.}
\label{fig:rectification_attention}
\end{figure}

Similar to the promotion attention module, the rectification attention module exists in the rectification subnetwork as shown in Fig.~\ref{fig:rectification_attention}. In the subnetwork, features from the five lateral connections of the feature extractor are decoded and as the input to each rectification branch in a multi-level manner. For each branch, the rectification attention is used to weight the object features $\text{F}_{\mathcal{R, O}}$ as follows:
\begin{equation}
\text{M}_{\mathcal{R,O}} = \text{F}_{\mathcal{R, O}} \otimes \text{A}_{\mathcal{R}} + \text{F}_{\mathcal{R, O}}.
\end{equation}
The generated features $\text{M}_{\mathcal{R,O}}$ is fed to a classifier, which is the same as the classifier in the promotion subnetwork. We denote the object outputs of rectification subnetwork as $\phi^{(i)}_{\mathcal{R,O}}(\pi^{(i)}_{\mathcal{R,O}}) \in [0, 1]^{H \times W \times 1}, i \in \{1, \dots, 5\}$, where $\pi^{(i)}_{\mathcal{R,O}}$ is the set of parameters of $\phi^{(i)}_{\mathcal{R,O}}$, consisting of the parameters of  decoding convolution layer and object branches. And the outputs of classifiers are expected to approximate the ground-truth masks of SOD. The minimizing optimization objective is as follows:
\begin{equation}
L_{\mathcal{R,O}} = \sum^5_{i=1} BCE(\phi^{(i)}_{\mathcal{R,O}}(\pi^{(i)}_{\mathcal{R,O}}), G).
\label{eq:rectification_object_network_loss}
\end{equation}

In addition, an additional regressor is added to the error-prone features $\text{F}_{\mathcal{R, E}}$. The regressor consists of two convolution layers with 128 kernels of $3 \times 3$ and $1 \times 1$, and one kernel of $1 \times 1$ followed by a Tanh operation. The outputs are the error-prone prediction of rectification subnetwork, denoted as $\phi^{(i)}_{\mathcal{R,E}}(\pi^{(i)}_{\mathcal{R,E}}) \in [-1, 1]^{H \times W \times 1}, i \in \{1,\dots, 5\}$, where $\pi^{(i)}_{\mathcal{R,E}}$ is the set of parameters of $\phi^{(i)}_{\mathcal{R,E}}$. The outputs of regressors aim to approximate the error maps of $\phi^{(i)}_{\mathcal{R,O}}(\pi^{(i)}_{\mathcal{R,O}})$ and the error map is defined as $G^{(i)}_{\mathcal{E}} = G - \phi^{(i)}_{\mathcal{R,O}}(\pi^{(i)}_{\mathcal{R,O}})$. Obviously, the value of $P^{(i)}_{\mathcal{E}}$ is in the range of $[-1, 1]$. In order to learning the error map, we drive predicted error map $P^{(i)}_{\mathcal{E}} = \phi^{(i)}_{\mathcal{R,E}}(\pi^{(i)}_{\mathcal{R,E}})$ to approach its ground truth $G^{(i)}_{\mathcal{E}}$ by minimizing the KL-divergence:
\begin{equation}
L_{\mathcal{R,E}} = \sum^5_{i=1} KL(N(G^{(i)}_{\mc{E}}) || N(P^{(i)}_{\mc{E}})),
\label{eq:rectification_error_network_loss}
\end{equation}
where $KL(\cdot || \cdot)$ means the KL-divergence with the following formulation:
\begin{equation}
KL(G||P) = \sum^{H \times W}_{i} G_{i}\mathrm{log} \frac{G_i}{P_i},
\end{equation}
where $P_{i}$ and $G_i$ are the $i$th pixel of the predicted error map $P^{(i)}_{\mathcal{E}}$ and the ground-truth error map of $G^{(i)}_{\mathcal{E}}$, respectively.
Also, $N(\cdot)$ in the above equation is a normalization operation, which casts the $G_{\mc{E}}$ and $P_{\mc{E}}$ into the range [0, 1]. In our method, we add 1 to the input and divide by 2 as the $N(\cdot)$ operation.

Through these operations, the rectification subnetwork provides the rectification attention and predicted error maps to the purificatory subnetwork, which drives PurNet to focus on the error-prone regions and rectify the wrong prediction.


\begin{figure}[t]
\centering
\includegraphics[width=1\columnwidth]{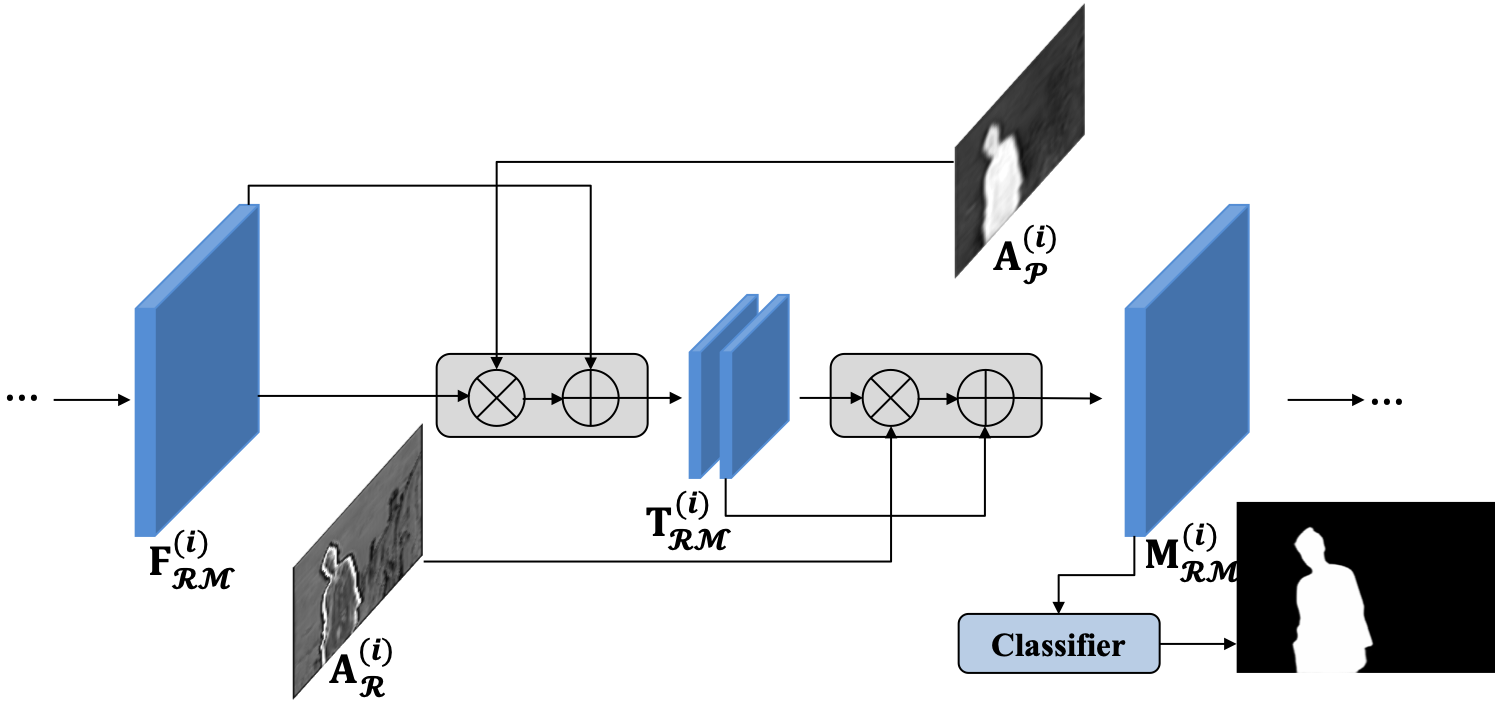}
\caption{The part structure of the purificatory subnetwork. The purificatory subnetwork integrates the promotion and rectification attentions by the purificatory mechanism.}
\label{fig:refinement_network}
\end{figure}

\subsection{Purificatory Subnetwork}
\subsubsection{Usage of Promotion and Rectification Attention}
Similar to the promotion subnetwork and rectification subnetwork, the purificatory subnetwork processes the features from feature extractor in a top-down manner, which can refine the SOD prediction progressively.

In our approach, the body of salient objects are first promoted with the help of the promotion attention, and then the error-prone regions of salient objects are rectified by the aid of then the rectification attention. Therefore, these two attentions are combined to purify the salient objects. The purificatory mechanism is integrated in the purificatory subnetwork, the structure of which is shown in~\ref{fig:refinement_network}. For $i$th decoding stage, the input features $\text{F}^{(i)}_{\mathcal{RM}} \in \mathbb{R}^{H \times W \times C}$ are firstly weighted by the promotion attention by the operation:
\begin{equation}
\text{T}^{(i)}_{\mathcal{RM}} = \text{F}^{(i)}_{\mathcal{RM}} \otimes \text{A}^{(i)}_{\mathcal{P}} + \text{F}^{(i)}_{\mathcal{RM}}.
\end{equation}
And then a convolution layer with 128 kernels of $3 \times 3$ are used to convolve the features to be $\text{T}'^{(i)}_{\mathcal{RM}}$. Next, the rectification attention is fed to weight the produced features as follows:
\begin{equation}
\text{M}^{(i)}_{\mathcal{RM}} = \text{T}'^{(i)}_{\mathcal{RM}} \otimes \text{A}^{(i)}_{\mathcal{R}} + \text{T}'^{(i)}_{\mathcal{RM}}.
\end{equation}
The generated features $\text{M}^{(i)}_{\mathcal{RM}}$ is input to a classifier, which is the same as the classifier in the promotion subnetwork with two convolution layers with 128 kernels of $3 \times 3$ and $1 \times 1$, and one kernel of $1 \times 1$ followed by a Sigmoid and upsampling operation.

We represent the outputs of the purificatory subnetwork as $\phi^{(i)}_{\mathcal{RM}}(\pi^{(i)}_{\mathcal{RM}}) \in [0, 1]^{H \times W \times 1}, i \in \{1, \dots, 5\}$, where $\pi^{(i)}_{\mathcal{RM}}$ is the set of parameters of $\phi^{(i)}_{\mathcal{RM}}$, consisting of the parameters of  decoding convolution layer and layers of each stage. And the outputs of classifiers are expected to approximate the ground truths of SOD. The loss is formed by the following operation:
\begin{equation}
L_{\mathcal{RM}} = \sum^5_{i=1} IBCE(\phi^{(i)}_{\mathcal{RM}}(\pi^{(i)}_{\mathcal{RM}}), G, P^{(i)}_{\mathcal{E}}),
\label{eq:refinement_network_loss}
\end{equation}
where $P^{(i)}_{\mathcal{E}} = \phi^{(i)}_{\mathcal{R,E}}(\pi^{(i)}_{\mathcal{R,E}})$ represents the error maps and $IBCE(\cdot, \cdot, \cdots)$ means the improved binary cross-entropy loss function with error map from the rectification subnetwork. We give the definition in Section~\ref{sec:improved_loss_function}. To provide more comprehensive visualization to prove the effectiveness of the proposed purificatory mechanism, we adopt the element-wise sum operation to combine these two features. Some examples without purificatory mechanism are shown in the fifth column of Fig.~\ref{fig:PM_results}.

\subsubsection{Improved Loss Function} \label{sec:improved_loss_function}
The predicted can be used to penalize the error-prone areas of the predicted saliency map in the purificatory subnetwork. By the extra constraints, the error-prone areas in the final prediction can be better refined. Toward this end, we propose to optimize the saliency maps to approximate the ground-truth masks of SOD by minimizing the improved binary cross-entropy loss (see Eq.~(\ref{eq:refinement_network_loss})). And the improved loss is defined as follows:
\begin{equation}
\begin{split}
& IBCE(P, G, E) = \\
& -\sum^{H \times W}_{i}(G_{i}\mathrm{log}P_i + (1 - G_{i})\mathrm{log}(1 - P_i)) \cdot (1 + \left|E_{i}\right|),
\end{split}
\end{equation}
where $E_i$ represents the $i$th pixel of predicted error maps $E$ and $\left|\cdot\right|$ indicates the absolute value operation. In our improved loss, the cross-entropy loss function at each pixel is weighted by the predicted error map, which penalizes the error-prone areas with bigger loss to rectify possible errors.

\section{Structural Similarity Loss}
Through the purificatory mechanism, different regions (\ie, simple regions and error-prone regions) of salient objects are processed and the performance is greatly improved. In addition to paying different attention to these indistinguishable regions, we also consider the structural constraints on complex regions as useful information for salient object detection. Toward this end, we propose a novel structural similarity loss (as shown in Fig.~\ref{fig:structural_loss_sp}) to constrain the region-level pairwise relationship between regions to calibrate the saliency values.

In general, current methods ({\eg, ~\cite{liu2018picanet,chen2018reverse,wang2017stagewise}) mainly adopt the binary cross-entropy loss function as the optimization objective, which is a pixel-level unary constraint for prediction by the formulation of Eq.~(\ref{eq:binary_cross_entropy}). However, Eq.~(\ref{eq:binary_cross_entropy}) only considers the relationship between each pixel and its corresponding ground-truth value, but does not take account of the relationship between different pixels or regions. In this manner, sometimes the saliency of whole local areas is completely detected incorrectly, which is caused by this problem lacking region-level relationship constraints.

\begin{figure}[t]
\centering
\includegraphics[width=1\columnwidth]{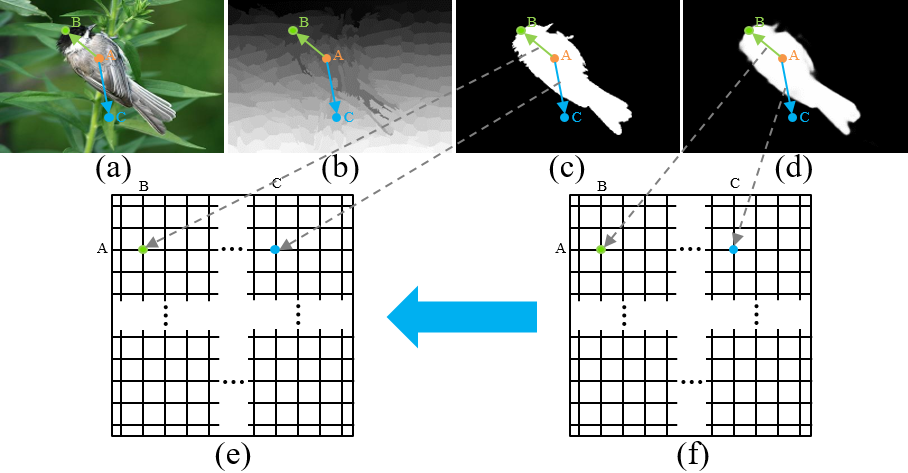}
\caption{Construction of the structural similarity matrix. (a) image, (b) super-pixel, (c)ground truth, (d) saliency map of our approach, (e)(f)structural matrices of the ground truth and saliency map.}
\label{fig:structural_loss_sp}
\end{figure}

To address this problem, we propose to model region-level pair-wise relationship as a supplement to the unary constraint and correct the probable errors. For the purpose of modeling the region-level relationship, we first construct a graph $\mathbb{G} = (\mathbb{V}, \mathbb{E})$ for each image, where $\mathbb{V}$ and $\mathbb{E}$ are the sets of nodes and directed edges. In the graph, each node represents a region $v_i \in \mathbb{V}, i = 1, \dots, N_{v}$ in images, where $N_{v}$ is the number of regions in the image. Regions in an image are easily generated by some existing methods~\cite{achanta2012slic,van2012seeds} and we adopt SLIC algorithm~\cite{achanta2012slic} to over-segment an RGB image into super-pixels as regions with $N_v = 256$. And the edge $v_i \rightarrow v_j$ from the region $v_i$ to the region $v_j$ represents the relation between these two regions. We apply the locations of super-pixels of RGB images to its corresponding ground truths and predicted saliency maps, and then we get the regions of the ground truths and predicted saliency maps.

For the ground truth and predicted saliency map of an RGB image, we define the saliency value of a region as the average of the sum of the saliency values of each pixel in this region, and the saliency value of $i$th super-pixel is denoted as $\mathcal{S}_{i} (i = 1, ..., N_{v})$. To model the relationship of regions, we use the subtraction of the saliency values between corresponding two nodes (\ie, $v_i$ and $v_j$) to represent the weight of each edge $v_i \rightarrow v_j$. Then, we construct structural matrix $\bl{M}$ to model the overall pair-wise relationship of an image as shown in Fig.~\ref{fig:structural_loss_sp}. The value in $i$th row and $j$th column of $\bl{M}$ represents the weight of the edge $v_i \rightarrow v_j$. In this manner, we can construct structural matrices for the ground-truth mask as $\bl{M}_{G}$ and predicted saliency map as $\bl{M}_{P}$ of every image. The ground truth and saliency map are expected to have the similar structure, so we drive $\bl{M}_{P}$ to become the structural similarity matrix of $\bl{M}_{G}$ by minimizing the KL-divergence:
\begin{equation}
SS(P, G) = D_{KL}(N(\bl{M}_{G}) || N(\bl{M}_{P})),
\end{equation}
where $N(\cdot)$ is a normalization operation as used in Eq.~(\ref{eq:rectification_error_network_loss}), and $P$ and $G$ means the predicted saliency map and ground-truth mask of an image, respectively. This loss function is named as the structural similarity loss (denoted as SSL).

In this work, we conduct the SSL on the outputs $\phi^{(i)}_{\mathcal{RM}}(\pi^{(i)}_{\mathcal{RM}})$ of each stage in the purificatory network, and the formulation of the overall structural similarity loss is as follows:
\begin{equation}
L_{\mathcal{SS}} = \sum^5_{i=1} SS(\phi^{(i)}_{\mathcal{RM}}(\pi^{(i)}_{\mathcal{RM}}), G).
\label{eq:rm_ssl}
\end{equation}

By taking the losses of Eqs.~(\ref{eq:promotion_network_loss}),~(\ref{eq:rectification_object_network_loss}),~(\ref{eq:rectification_error_network_loss}),~(\ref{eq:refinement_network_loss}) and~(\ref{eq:rm_ssl}), the overall learning objective can be formulated as follows:
\begin{equation}
\min_{\mathbb{P}} L_{\mathcal{P}} + L_{\mathcal{R,O}} + L_{\mathcal{R,E}} + L_{\mathcal{RM}} + L_{\mathcal{SS}},
\label{eq:overall_loss}
\end{equation}
where $\mathbb{P}$ is the set of $\{\pi_i, \pi^{(i)}_\mathcal{P}, \pi^{(i)}_\mathcal{R,O}, \pi^{(i)}_\mathcal{R,E}, \pi^{(i)}_\mathcal{RM}\}^5_{i=1}$ for convenience of presentation.



\begin{table*}[t]
\centering
\caption{Performance on six benchmark datasets. Smaller MAE, larger $F^{w}_{\beta}$ and $F_{\beta}$ correspond to better performance. The best results of different backbones are in \textbf{\color{blue}{blue}} and \textbf{\color{red}{red}} fonts. ``-" means the results cannot be obtained and ``$^\dagger$'' means the results are post-processed by dense conditional random field (CRF)~\cite{krahenbuhl2011efficient}. Note that the backbone of PAGRN is VGG-19~\cite{simonyan2015very} and the one of R3Net is ResNeXt-101~\cite{xie2017aggregated}.
MK: MSRA10K~\cite{cheng2015global}, DUTS: DUTS-TR~\cite{wang2017learning}, MB: MSRA-B~\cite{liu2010learning}.}
\setlength{\tabcolsep}{1.15mm}{
\begin{tabular}{c|c|c c c c c c c c c c c c c c c c c c }
\hline
\multirow{2}*{Models} & Training  & \multicolumn{3}{ c|}{ECSSD} & \multicolumn{3}{c|}{DUT-OMRON} & \multicolumn{3}{c|}{PASCAL-S} & \multicolumn{3}{c|}{HKU-IS} & \multicolumn{3}{c|}{DUTS-TE} & \multicolumn{3}{c}{XPIE}\\
\cline{3-20}
 & Dataset & MAE & $F^{w}_{\beta}$ & \multicolumn{1}{ c|}{$F_{\beta}$} & MAE & $F^{w}_{\beta}$ & \multicolumn{1}{ c|}{$F_{\beta}$} & MAE & $F^{w}_{\beta}$ & \multicolumn{1}{ c|}{$F_{\beta}$}
& MAE & $F^{w}_{\beta}$ & \multicolumn{1}{ c|}{$F_{\beta}$} & MAE & $F^{w}_{\beta}$ & \multicolumn{1}{ c|}{$F_{\beta}$} & MAE & $F^{w}_{\beta}$ & \multicolumn{1}{ c}{$F_{\beta}$} \\
\hline
\multicolumn{20}{c}{VGG-16 backbone~\cite{simonyan2015very}} \\
\hline

KSR~\cite{wang2016kernelized} & MB
            & 0.132 & 0.633 & 0.810 & 0.131 & 0.486 & 0.625 & 0.157 & 0.569 & 0.773 & 0.120 & 0.586 & 0.773 & - & - & - & - & - & - \\
HDHF~\cite{li2016visual} & MB
            & 0.105 & 0.705 & 0.834 & 0.092 & 0.565 & 0.681 & 0.147 & 0.586 & 0.761 & 0.129 & 0.564 & 0.812 & - & - & - & - & - & - \\
ELD~\cite{lee2016deep} &  MK
            & 0.078 & 0.786 & 0.829 & 0.091 & 0.596 & 0.636 & 0.124 & 0.669 & 0.746 & 0.063 & 0.780 & 0.827 & 0.092 & 0.608 & 0.647 & 0.085 & 0.698 & 0.746 \\
UCF~\cite{zhang2017learning}  & MK
            & 0.069 & 0.807 & 0.865 & 0.120 & 0.574 & 0.649 & 0.116 & 0.696 & 0.776 & 0.062 & 0.779 & 0.838 & 0.112 & 0.596 & 0.670 & 0.095 & 0.693 & 0.773 \\
NLDF~\cite{luo2017non} & MB
            & 0.063 & 0.839 & 0.892 & 0.080 & 0.634 & 0.715 & 0.101 & 0.737 & 0.806 & 0.048 & 0.838 & 0.884 & 0.065 & 0.710 & 0.762 & 0.068 & 0.762 & 0.825 \\
Amulet~\cite{zhang2017amulet} & MK
            & 0.059 & 0.840 & 0.882 & 0.098 & 0.626 & 0.673 & 0.099 & 0.736 & 0.795 & 0.051 & 0.817 & 0.853 & 0.085 & 0.658 & 0.705 & 0.074 & 0.743 & 0.796 \\
FSN~\cite{chen2017look} &  MK
            & 0.053 & 0.862 & 0.889 & 0.066 & 0.694 & 0.733 & 0.095 & 0.751 & 0.804 & 0.044 & 0.845 & 0.869 & 0.069 & 0.692 & 0.728 & 0.066 & 0.762 & 0.812 \\
C2SNet~\cite{li2018contour} & MK
            & 0.057 & 0.844 & 0.878 & 0.079 & 0.643 & 0.693 & 0.086 & 0.764 & 0.805 & 0.050 & 0.823 & 0.854 & 0.065 & 0.705 & 0.740 & 0.066 & 0.764 & 0.807 \\
RA~\cite{chen2018reverse} &MB
            & 0.056 & 0.857 & 0.901 & 0.062 & 0.695 & 0.736 & 0.105 & 0.734 & 0.811 & 0.045 & 0.843 & 0.881 & 0.059 & 0.740 & 0.772 & 0.067 & 0.776 & 0.836 \\
PAGRN~\cite{zhang2018progressive} & DUTS
            & 0.061 & 0.834 & 0.912 & 0.071 & 0.622 & 0.740 & 0.094 & 0.733 & 0.831 & 0.048 & 0.820 & 0.896 & 0.055 & 0.724 & 0.804 & - & - & -\\

RFCN~\cite{wang2019salient} & MK
            & 0.067 & 0.824 & 0.883 & 0.077 & 0.635 & 0.700 & 0.106 & 0.720 & 0.802 & 0.055 & 0.803 & 0.864 & 0.074 & 0.663 & 0.731 & 0.073 & 0.736 & 0.809 \\
DSS$^\dagger$~\cite{hou2019deeply} & MB
                    & 0.052 & 0.872 & 0.918 & 0.063 & 0.697 & \textbf{\color{blue}{0.775}} & 0.098 & 0.756 & 0.833 & 0.040 & 0.867 & \textbf{\color{blue}{0.904}} & 0.056 & 0.755 & 0.810 & 0.065 & 0.784 & 0.849 \\
MLM~\cite{wu2019mutual} & DUTS
            & 0.045 & 0.871 & 0.897 & 0.064 & 0.681 & 0.719 & 0.077 & 0.778 & 0.813 & 0.039 & 0.859 & 0.882 & 0.049 & 0.761 & 0.776 & - & - & -\\
AFNet~\cite{feng2019attentive} & DUTS
                    &  0.042 & 0.886 & 0.916 & 0.057 & 0.717 & 0.761 & \textbf{\color{blue}{0.073}} & 0.797 & 0.839 & \textbf{\color{blue}{0.036}}  & 0.869 & 0.895 & 0.046  & 0.785  & 0.807 & 0.047  &  0.822  & 0.859  \\

\hline
\textbf{Ours} &  MK
            & 0.042 & 0.887 & 0.914
            & 0.064 & 0.708 & 0.743
            & 0.080 & 0.779 & 0.830
            & 0.042 & 0.852 & 0.885
            & 0.052 & 0.768 & 0.792
            & 0.053 & 0.808 & 0.851 \\

\textbf{Ours} &   DUTS &
            \textbf{\color{blue}{0.040}} & \textbf{\color{blue}{0.892}} & \textbf{\color{blue}{0.920}} &
            \textbf{\color{blue}{0.054}} & \textbf{\color{blue}{0.730}} & 0.768 &
            0.076 & \textbf{\color{blue}{0.798}} & \textbf{\color{blue}{0.841}} &
            \textbf{\color{blue}{0.036}} & \textbf{\color{blue}{0.870}} & 0.894 &
            \textbf{\color{blue}{0.043}} &\textbf{\color{blue}{0.797}}&\textbf{\color{blue}{0.816}} &
            \textbf{\color{blue}{0.044}} & \textbf{\color{blue}{0.830}} & \textbf{\color{blue}{0.868}} \\

\hline
\multicolumn{20}{c}{ResNet-50 backbone~\cite{he2016deep}} \\
\hline

SRM~\cite{wang2017stagewise} & DUTS
            & 0.054 & 0.853 & 0.902 & 0.069 & 0.658 & 0.727 & 0.086 & 0.759 & 0.820 & 0.046 & 0.835 & 0.882 & 0.059 & 0.722 & 0.771 & 0.057 & 0.783 & 0.841 \\
Picanet~\cite{liu2018picanet} & DUTS
            & 0.047 & 0.866 & 0.902 & 0.065 & 0.695 & 0.736 & 0.077 & 0.778 & 0.826 & 0.043 & 0.840 & 0.878 & 0.051 & 0.755 & 0.778 & 0.052 & 0.799 & 0.843 \\
R3$^\dagger$~\cite{deng2018r3net} &  MK
            & 0.040 & 0.902 & 0.924 & 0.063 & 0.728 & 0.768 & 0.095 & 0.760 & 0.834 & 0.036 & 0.877 & 0.902 & 0.057 & 0.765 & 0.805 & 0.058 & 0.805 & 0.854 \\
DGRL~\cite{wang2018detect} & DUTS
            & 0.043 & 0.883 & 0.910 & 0.063 & 0.697 & 0.730 & 0.076 & 0.788 & 0.826 & 0.037 & 0.865 & 0.888 & 0.051 & 0.760 & 0.781 & 0.048 & 0.818& 0.859 \\

ICTBI~\cite{wang2019iterative} &  DUTS
            &  0.041 & 0.881 & 0.909 & 0.061 & 0.730 & 0.758 & 0.071 & 0.788 & 0.826 & 0.038 & 0.856 & 0.890 & 0.048 & 0.762 & 0.797 & - & - & -\\

            \hline

\textbf{Ours} &  MK
            & 0.038 & 0.896 & 0.917
            & 0.063 & 0.728 & 0.754
            & 0.075 & 0.801 & 0.842
            & 0.036 & 0.871 & 0.892
            & 0.047 & 0.780 & 0.802
            & 0.046 & 0.823 & 0.860 \\

\textbf{Ours}  & DUTS
            & \textbf{\color{red}{0.035}} & \textbf{\color{red}{0.907}} & \textbf{\color{red}{0.928}} & \textbf{\color{red}{0.051}} & \textbf{\color{red}{0.747}} & \textbf{\color{red}{0.776}} & \textbf{\color{red}{0.070}} & \textbf{\color{red}{0.805}} & \textbf{\color{red}{0.847}} & \textbf{\color{red}{0.031}} & \textbf{\color{red}{0.889}} & \textbf{\color{red}{0.904}} & \textbf{\color{red}{0.039}} & \textbf{\color{red}{0.817}} & \textbf{\color{red}{0.829}} & \textbf{\color{red}{0.041}} & \textbf{\color{red}{0.843}} & \textbf{\color{red}{0.876}} \\
\hline
\end{tabular}}
\label{tab:result-all}
\end{table*}

\section{Experiments}
\subsection{Experimental Setup}
\subsubsection{Datasets}
To evaluate the performance of our method, we conduct experiments on six benchmark datasets~\cite{yan2013hierarchical,yang2013saliency,li2015visual,wang2017learning,xia2017and}. Details of these datasets are briefly described as follows: ECSSD~\cite{yan2013hierarchical} consists of 1,000 images with complex and semantically meaningful objects. DUT-OMRON~\cite{yang2013saliency} has 5,168 complex images that are downsampled to a maximal side length of 400 pixels. PASCAL-S~\cite{li2014secrets} includes 850 natural images that are pre-segmented into objects or regions with salient object annotation by eye-tracking test of 8 subjects. HKU-IS~\cite{li2015visual} contains 4,447 images which usually contain multiple disconnected salient objects or salient objects that touch image boundaries. DUTS~\cite{wang2017learning} is a large scale dataset containing 10,533 training images (named as DUTS-TR) and 5019 test images(named as DUTS-TE). The images are challenging with salient objects that occupy various locations and scales as well as complex background. XPIE~\cite{xia2017and} is also a large dataset that has 10,000 images covering a variety of simple and complex scenes with various salient objects.

\subsubsection{Evaluation Metrics}
We choose mean absolute error (MAE), weighted F-measure score ($F^w_{\beta}$)~\cite{margolin2014evaluate}, F-measure score ($F_{\beta}$), and F-measure curve to evaluate our method. MAE is the average pixel-wise absolute difference between ground-truth masks and estimated saliency maps. In computing F$_{\beta}$, we normalize the predicted saliency maps into the range [0, 255] and binarize the saliency maps with a threshold sliding from 0 to 255 to compare the binary maps with ground-truth masks. At each threshold, Precision and Recall can be computed. F$_{\beta}$ is computed as:
\begin{equation}\label{eq:Fbeta}
F_{\beta} = \frac{(1 + \beta^2) \cdot Precision \cdot Recall}{\beta^2 \cdot Precision + Recall},
\end{equation}
where $\beta^2$ is set to 0.3 to emphasize more on Precision than Recall as suggested in \cite{achanta2009frequency}. Then we can plot F-measure curve based on all the binary maps over all saliency maps in a given dataset. We report F$_{\beta}$ using an adaptive threshold for generating binary a saliency map and the threshold is computed as twice the mean of a saliency map. In addition, $F^w_\beta$ is used to evaluate the overall performance (more details can be found in ~\cite{margolin2014evaluate}).

\subsubsection{Training and Inference.}
We train the networks in three stages and the training steps as follows: \textcircled{\small{1}} we first train the feature extractor and purificatory subnetwork with $L_{\mc{RM}}$ and $L_{\mc{SS}}$; \textcircled{\small{2}} we fix the purificatory subnetwork then train the promotion rectification subnetworks with $L_{\mc{P}}$, $L_{\mc{R,O}}$ and $L_{\mc{RE}}$; \textcircled{\small{3}} Then we train the whole network with the overall loss in Eq.~(\ref{eq:overall_loss}).

We use standard stochastic gradient descent algorithm to train our network end-to-end by optimizing the learning object in Eq.~(\ref{eq:overall_loss}). In the optimization process, the parameters of feature extractor is initialized by the pre-trained backbone model \cite{he2016deep}, whose learning rate is set to $1 \times 10^{-3}$ with a weight decay of $5 \times 10^{-4}$ and momentum of 0.9. And the learning rate of rest layers are set to 10 times larger. Besides, we employ the ``poly'' learning rate policy for all experiments similar to \cite{liu2015parsenet}.

We train our network with ResNet-50~\cite{he2016deep} by utilizing the training set of DUTS-TR dataset \cite{wang2017learning} as used in~\cite{wang2018detect,zhang2018progressive,liu2018picanet,wang2017stagewise} and MSRA10K~\cite{deng2018r3net}. The training images are resized to the resolution of $320 \times 320$ for faster training, and applied horizontal flipping. For a more comprehensive demonstration and fairer comparison, we also use VGG-16~\cite{simonyan2015very} as the backbone of our method instead of ResNet-50~\cite{he2016deep}, and train the new network without changing other settings. The training process takes about 20 hours and converges after 500k iterations (20k iterations for stage \textcircled{\small{1}}, 50k iterations for stage \textcircled{\small{2}}  and 200k iterations for stage \textcircled{\small{3}}) with mini-batch of size 8 on a single NVIDA TITAN Xp GPU. During inference, the proposed network removes all the losses, and one image is directly fed into the network to produce the saliency map at the output of first stage in the purificatory network. And the network runs at about 27fps on a single NVIDIA 1080Ti GPU for inference.

\subsection{Comparisons with the State-of-the-art}
\begin{figure*}[t]
\centering
\includegraphics[width=1\textwidth]{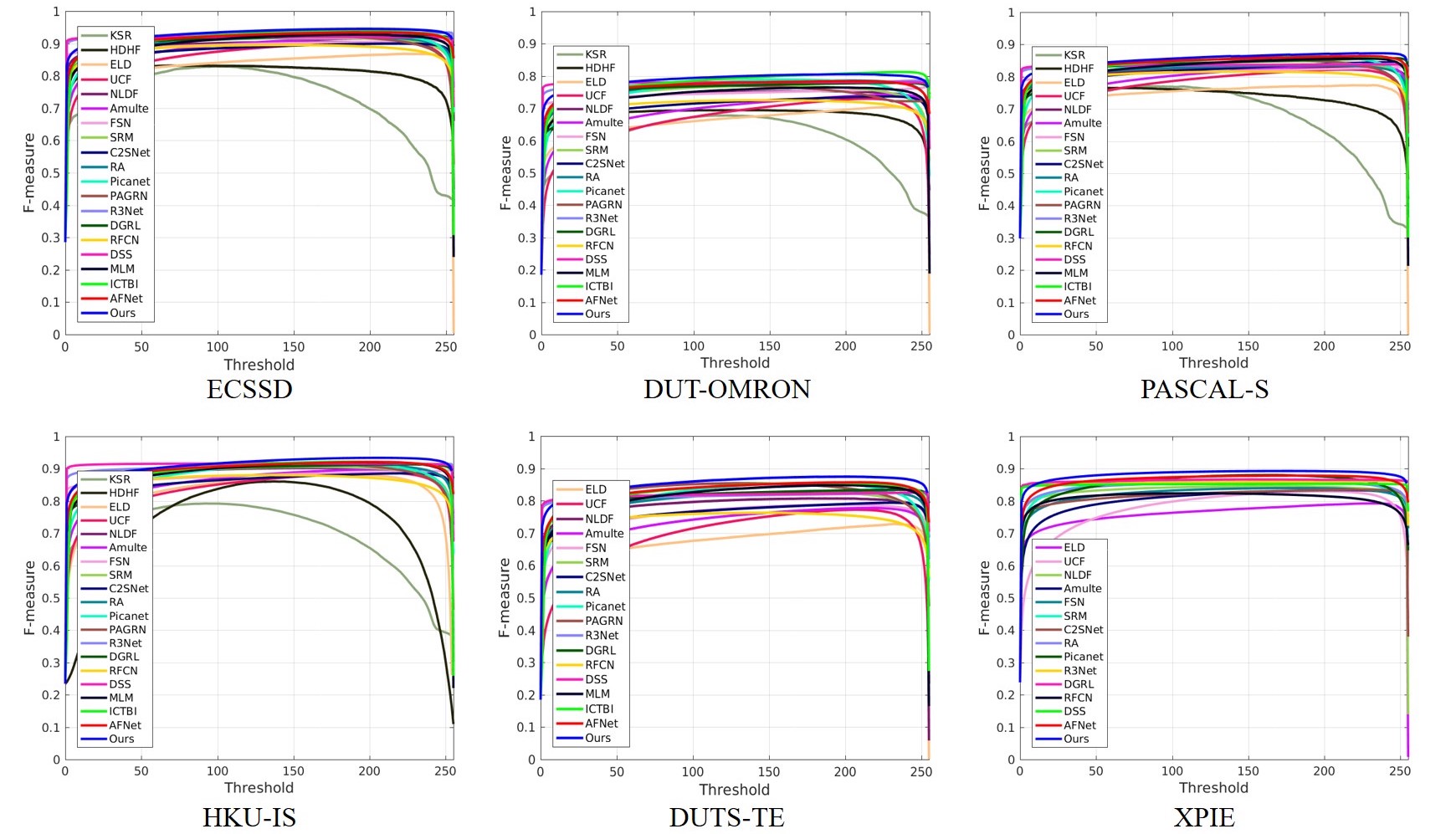}
\caption{The F-measure curves of 19 state-of-the-arts and our approach are listed across six benchmark datasets.}
\label{fig:prfmeasure}
\end{figure*}

\begin{figure*}[t]
\centering
\includegraphics[width=1\textwidth]{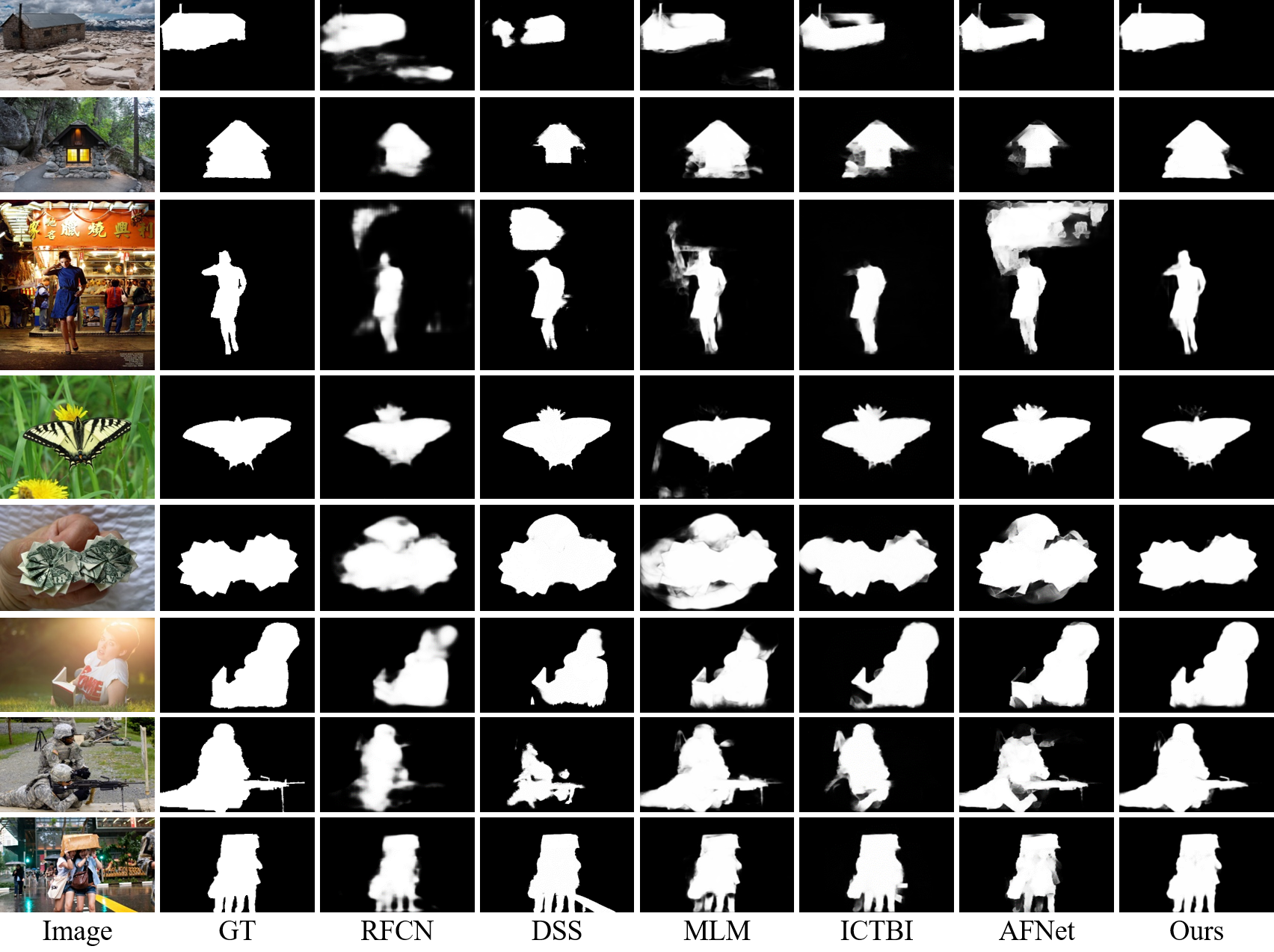}
\caption{Qualitative comparisons of the state-of-the-art algorithms and our approach. GT means ground-truth masks of salient objects.}
\label{fig:result-all}
\end{figure*}

We compare our approach denoted as with 19 state-of-the-art methods, including KSR~\cite{wang2016kernelized}, HDHF~\cite{li2016visual}, ELD \cite{lee2016deep}, UCF \cite{zhang2017learning}, NLDF \cite{luo2017non}, Amulet \cite{zhang2017amulet}, FSN \cite{chen2017look}, SRM \cite{wang2017stagewise}, C2SNet~\cite{li2018contour}, RA~\cite{chen2018reverse}, Picanet~\cite{liu2018picanet}, PAGRN \cite{zhang2018progressive}, R3Net~\cite{deng2018r3net},  DGRL~\cite{wang2018detect}, RFCN~\cite{wang2019salient}, DSS~\cite{hou2019deeply}, MLM~\cite{wu2019mutual}, ICTBI~\cite{wang2019iterative} and AFNet~\cite{feng2019attentive}. We obtain the saliency maps of different methods from the authors or the deployment codes provided by the authors for fair comparison.

\subsubsection{Quantitative Evaluation}
We evaluate 19 state-of-the-art SOD methods and our method on six benchmark datasets with different backbones and training sets, and the results are listed in Tab.~\ref{tab:result-all}. We can see that the proposed method consistently outperform other methods across all the six datasets, especially DUTS-TE and XPIE.

When training with ResNet-50, our method is noticeably improved from 0.765 to 0.817 on DUTS-TE and from 0.818 to 0.843 on XPIE compared to the second best results as for $F^w_{\beta}$. Also, it is worth noting that $F_\beta$ of our method is significantly better compared with the second best results on DUTS-TE (0.829 against 0.805) and XPIE (0.876 against 0.859). As for MAE, our method has obvious advantages compared with other state-of-the-art algorithms on six datasets. Similarly, PurNet has an analogous and obvious improvement when training our network with VGG-16 as backbone. The advantages on these datasets confirm that our proposed purificatory mechanism and similarity structural loss can achieve great performance with different backbones.


For overall comparisons, F-measure curves of different methods are displayed in Fig.~\ref{fig:prfmeasure}. We can observe that the F-measure curves of our approach are consistently higher than other state-of-the-art methods. These observations present the efficiency and robustness of our purificatory network across various challenging datasets, which indicates that the perspective of purificatory mechanism for the problem of SOD is useful. Note that the results of DSS, RA on HKU-IS~\cite{li2015visual} are only conducted on the test set.

\begin{table*}[t]
\centering
\caption{Performance of differents setting of the proposed method. PurNet is the proposed method. Meanings of other abbreviations are as follows: RM: purificatory Network, PA: Promotion Attention Network, RA: Rectification Attention Network, SSL: Structural Similarity Loss.}
\setlength{\tabcolsep}{3.1mm}{
\renewcommand\arraystretch{1.0}
\begin{tabular}{c|c c c c|ccccccccc}
\hline
& \multirow{2}*{RM} & \multirow{2}*{PA} & \multirow{2}*{RA} & \multirow{2}*{SSL} & \multicolumn{3}{c|}{ECSSD} & \multicolumn{3}{c|}{DUT-OMRON} & \multicolumn{3}{c}{PASCAL-S} \\
\cline{6-14}
& & & &  & MAE & $F^{w}_{\beta}$ & \multicolumn{1}{c|}{$F_{\beta}$} & MAE & $F^{w}_{\beta}$ & \multicolumn{1}{c|}{$F_{\beta}$} & MAE & $F^{w}_{\beta}$ & $F_{\beta}$ \\
\hline
Baseline & \checkmark & & &
            & 0.046 & 0.864 & 0.895 & 0.064 & 0.700 & 0.725 & 0.077 & 0.776 & 0.820 \\
Baseline + PA & \checkmark & \checkmark & &
            &  0.044 & 0.877 & 0.919 & 0.055 & 0.719 & 0.760 & 0.079 & 0.778 & 0.838\\
Baseline + RA & \checkmark &  & \checkmark &
                    & 0.043 & 0.878 & 0.917 & 0.053 & 0.725 & 0.765 & 0.076 & 0.781 & 0.838 \\
Baseline + PM & \checkmark &\checkmark  &\checkmark  &
			& 0.039 & 0.892 & 0.924 & 0.055 & 0.734 & 0.768 & 0.071 & 0.798 & \textbf{\underline{0.848}} \\
Baseline + SSL & \checkmark &  &  & \checkmark
		 	& 0.043 & 0.880 & 0.917 & 0.057 & 0.716 & 0.760 & 0.074 & 0.786 & 0.838 \\
            \hline
\textbf{PurNet} & \checkmark & \checkmark &\checkmark &\checkmark
            & \textbf{\underline{0.035}} & \textbf{\underline{0.907}} & \textbf{\underline{0.928}} & \textbf{\underline{0.051}} & \textbf{\underline{0.747}} & \textbf{\underline{0.776}} & \textbf{\underline{0.070}} & \textbf{\underline{0.805}} & 0.847  \\
\hline
\end{tabular}}
\label{tab:different_settings}
\end{table*}

\begin{table}[t]
\centering
\caption{Comparisons of each side-outputs and their fusion. S$i (i=1,\dots,5)$ means $t$th side-output and Fusion means the average of S1 to S5.}
\setlength{\tabcolsep}{1.15mm}{
\renewcommand\arraystretch{1.0}
\begin{tabular}{c|ccccccccc}
\hline
\multirow{2}*{} & \multicolumn{3}{c|}{ECSSD} & \multicolumn{3}{c|}{DUT-OMRON} & \multicolumn{3}{c}{PASCAL-S} \\
\cline{2-10}
&  MAE & $F^{w}_{\beta}$ & \multicolumn{1}{c|}{$F_{\beta}$}  & MAE & $F^{w}_{\beta}$ & \multicolumn{1}{c|}{$F_{\beta}$} & MAE & $F^{w}_{\beta}$ & \multicolumn{1}{c}{$F_{\beta}$} \\
\hline
S5
            & 0.043 & 0.878 & 0.899 & 0.057 & 0.712 & 0.741 & 0.077 & 0.778 & 0.818 \\
S4
            & 0.043 & 0.880 & 0.899 & 0.057 & 0.715 & 0.742 & 0.076 & 0.781 & 0.819 \\
S3
            & 0.037 & 0.900 & 0.920 & 0.053 & 0.738 & 0.765 & 0.071 & 0.799 & 0.841 \\
S2
            & \textbf{\underline{0.035}} & 0.906 & 0.927 & \textbf{\underline{0.051}} & 0.746 & 0.775 & \textbf{\underline{0.070}} & 0.804 & 0.846 \\
S1
            & \textbf{\underline{0.035}} & \textbf{\underline{0.907}} & \textbf{\underline{0.928}} & \textbf{\underline{0.051}} & \textbf{\underline{0.747}} & \textbf{\underline{0.776}} & \textbf{\underline{0.070}} & \textbf{\underline{0.805}} & \textbf{\underline{0.847}} \\
Fusion
            &  0.038 & 0.894 & 0.918 & 0.054 & 0.733 & 0.758 & 0.073 & 0.794 & 0.837\\
\hline
\end{tabular}}
\label{tab:result-side-output}
\end{table}

\begin{table*}[t]
\centering
\caption{Performance compared with the latest methods published in 2019 and 2020. Smaller MAE, larger S-Measure and E-Measure correspond to better performance. The best and second results are in \textbf{\color{red}{red}} and \textbf{\color{blue}{blue}} fonts.}
\setlength{\tabcolsep}{1.15mm}{

\begin{tabular}{c|c|c c c c c c c c c c c c }
\hline
\multirow{2}*{Models} & \multirow{2}*{Year}
& \multicolumn{3}{ c|}{ECSSD}
& \multicolumn{3}{c|}{DUT-OMRON}
& \multicolumn{3}{c|}{HKU-IS}
& \multicolumn{3}{c}{DUTS-TE}\\
\cline{3-14}
&  & MAE & S-measure & \multicolumn{1}{ c|}{E-measure} &
     MAE & S-measure & \multicolumn{1}{ c|}{E-measure} &
     MAE & S-measure & \multicolumn{1}{ c|}{E-measure} &
     MAE & S-measure & \multicolumn{1}{ c}{E-measure} \\
\hline
PAGENet~\cite{wangwenguan2019salient} & 2019
            & 0.040 & 0.921 & 0.911
            & 0.061 & 0.851 & 0.822
            & 0.050 & 0.873 & 0.851
            & 0.034 & 0.941 & 0.902  \\
CPD-R~\cite{wu2019cascaded} & 2019
            & 0.037 & 0.925 & 0.918
            & 0.056 & 0.866 & 0.825
            & 0.043 & 0.887 & 0.869
            & 0.034 & 0.944 & 0.905  \\
BASNet~\cite{qin2019basnet} & 2019
            & 0.037 & 0.921 & 0.916
            & 0.056 & \textbf{\color{red}{0.869}} & 0.836
            & 0.048 & 0.884 & 0.866
            & 0.032 & 0.946 & 0.909 \\
ITSDNet~\cite{zhou2020interactive} & 2020
            & \textbf{\color{blue}{0.034}} & \textbf{\color{red}{0.927}} & \textbf{\color{blue}{0.925}}
            & 0.061 & 0.863 & \textbf{\color{blue}{0.840}}
            & 0.041 & 0.895 & \textbf{\color{blue}{0.885}}
            & 0.031 & \textbf{\color{blue}{0.952}} & 0.917 \\
MINet~\cite{pang2020multi} & 2020
            & \textbf{\color{red}{0.033}} & \textbf{\color{red}{0.927}} & \textbf{\color{blue}{0.925}}
            & \textbf{\color{blue}{0.055}} & 0.865 & 0.833
            & \textbf{\color{red}{0.037}} & \textbf{\color{red}{0.898}} & 0.884
            & \textbf{\color{red}{0.029}} & \textbf{\color{red}{0.953}} & \textbf{\color{blue}{0.919}} \\
GCPANet~\cite{chen2020global} & 2020
            & 0.035 & 0.920 & \textbf{\color{red}{0.927}}
            & 0.056 & 0.860 & 0.839
            & \textbf{\color{blue}{0.038}} & 0.891 & \textbf{\color{red}{0.891}}
            & 0.031 & 0.949 & \textbf{\color{red}{0.920}} \\

            \hline
\textbf{Ours} & -
            & 0.035 & \textbf{\color{blue}{0.925}} & \textbf{\color{blue}{0.925}}
            & \textbf{\color{red}{0.051}} & \textbf{\color{blue}{0.868}} & \textbf{\color{red}{0.841}}
            & 0.039 & \textbf{\color{blue}{0.897}} & 0.880
            & \textbf{\color{blue}{0.031}} & 0.950 & 0.917 \\
\hline
\end{tabular}}
\label{tab:result-newadd}
\end{table*}

\textbf{Qualitative Evaluation.}
Some examples of saliency maps generated by our approach and other state-of-the-art algorithms are shown in Fig.~\ref{fig:result-all}. We can see that salient objects can pop-out with accurate location and details by the proposed method. From the row of 1 to 3 in Fig.~\ref{fig:result-all}, we can find that many methods usually can't locate the salient objects roughly. In our method, the salient objects are located with the help of effective promotion attention. In addition, lots of methods often mistakenly segment the details of salient objects.  We think the reason for this error is that most existing methods usually lack the constraints of error-prone areas. From the row of 4 to 6 in Fig.~\ref{fig:result-all}, we can observe that our method achieves better performance, which indicates the ability of processing the fine structures and rectifying errors.
More examples of complex scenes are shown in the row of 7 and 8, we can observe that the proposed method also obtains the impressive results. These observations indicate that addressing SOD from the perspective of purificatory mechanism and region-level pair-wise constraints is effective.

\subsection{Ablation Studies}

To validate the effectiveness of different components of the proposed method, we conduct several experiments on the benchmark datasets to compare the performance variations of our methods with different experimental settings.

\subsubsection{Effectiveness of the Purificatory Mechanism}
To investigate the effectiveness of the proposed purificatory mechanism, we conduct ablation experiments and introduce four different models for comparisons. The first setting is only the feature extractor and purificatory subnetwork, which is regarded as ``Baseline''. To explore the respective effectiveness of promotion attention and rectification attention, we conduct the second and third model by adding the promotion subnetwork (denoted as ``Baseline + PA'') and rectification subnetwork  (denoted as ``Baseline + RA''), respectively. In addition, we combine the two attention mechanisms (\ie, purificatory mechanism) with the purificatory network as the fourth models, which is named as ``Baseline + PM''. We also list the proposed method with the purificatory mechanism and structural similarity loss as ``\textbf{PurNet}''.

The comparison results of above mentioned models are listed in Tab.~\ref{tab:different_settings}. We can observe that the promotion attention and rectification attention greatly improve the performance compared with ``Baseline'', which indicates the usefulness of the two attention mechanisms for SOD. In addition, we can find that ``Baseline + RA'' has better performance improvement than ``Baseline + PA'', which implies that the rectification of some error-prone areas is important to SOD. Moreover, a better performance has been achieved through the combination of the two attentions (\ie, purificatory mechanism), which verifies the compatibility of the two attentions and effectiveness of the purificatory mechanism.

\subsubsection{Effectiveness of the Structural Similarity Loss}
To investigate the effectiveness of the proposed novel  structural similarity loss (SSL), we conduct another experiments by only combining the loss with ``Baseline'' and this model is named as ``Baseline + SSL''. As listed in Tab.~\ref{tab:different_settings}, we can observe a remarkable improvement brought by SSL by comparing ``Baseline'' and ``Baseline + SSL''. The result shows that the loss plays an important role in the SOD task. In addition, by comparing ``Baseline + PM'' and ``\textbf{PurNet}'', we can find that SSL is still useful even when the results is advanced.

\subsubsection{Performance of Each Side-output}
In order to explore how to obtain the best prediction of the proposed network, we conduct an additional experiment to compare the performance of each side-outputs and fusion in the purificatory subnetwork. As listed in Tab.~\ref{tab:result-side-output}, we can see that the performance of last three side-outputs (\ie, third, fourth and fifth side-output) is consistently worse than the one of the first two side-outputs (\ie, first and second side-output). And the performance of fusion is lower than first, second and third side-output. The comparisons indicate the process of generating saliency maps in our network is progressively refined from the higher layer to the lower layer. Thus, we choose the first side-output as the results during inference.

\subsubsection{Compared with the latest methods published in 2019 and 2020 with new evaluation metrics}
It is worth noting that since our method was submitted in 2019, many new works on salient object detection have been published since then. For a more fair comparison, we discussed and added several methods which are similar to our experimental settings for comparison. In addition, besides mean absolute error (MAE), we used two new evaluation metrics to verify the performance of the method from multiple perspectives, namely S-measure~\cite{fan2017structure} and E-measure~\cite{fan2018enhanced}. The experimental results are shown in Tab.~\ref{tab:result-newadd}. It can be seen that compared with the method published in 2019, our method has obvious advantages, and compared with the latest method published in 2020, our performance is still competitive.

\section{Conclusion}
In this paper, we rethink the two difficulties that hinder the development of salient object detection. The difficulties consists of indistinguishable regions and complex structures. To solve these two issues, we propose the purificatory network with structural similarity loss. In this network, we introduce the promotion attention to improve the localization ability and semantic information for salient regions, which guides the network to focus on salient regions. We also propose the rectification subnetwork to provide the rectification attention for rectifying the errors. The two attentions are combined to form the purificatory mechanism to improve the promotable regions and rectifiable regions for purifying salient objects. Moreover, we also propose a novel region-level pair-wise structural similarity loss, which models and constrains the relationships between pair-wise regions. This loss can be used to be as a supplement to the unary constraint. Extensive experiments on six benchmark datasets have validated the effectiveness of the proposed approach.

\section*{ACKNOWLEDGMENT}

This work was supported in part by the National Natural Science Foundation of China under Grant 61922006, and Grant 62088102.

\bibliographystyle{IEEEtran}
\bibliography{RefPurNet}
\begin{IEEEbiography}[{\includegraphics[width=0.8in,height=1in,clip,keepaspectratio]{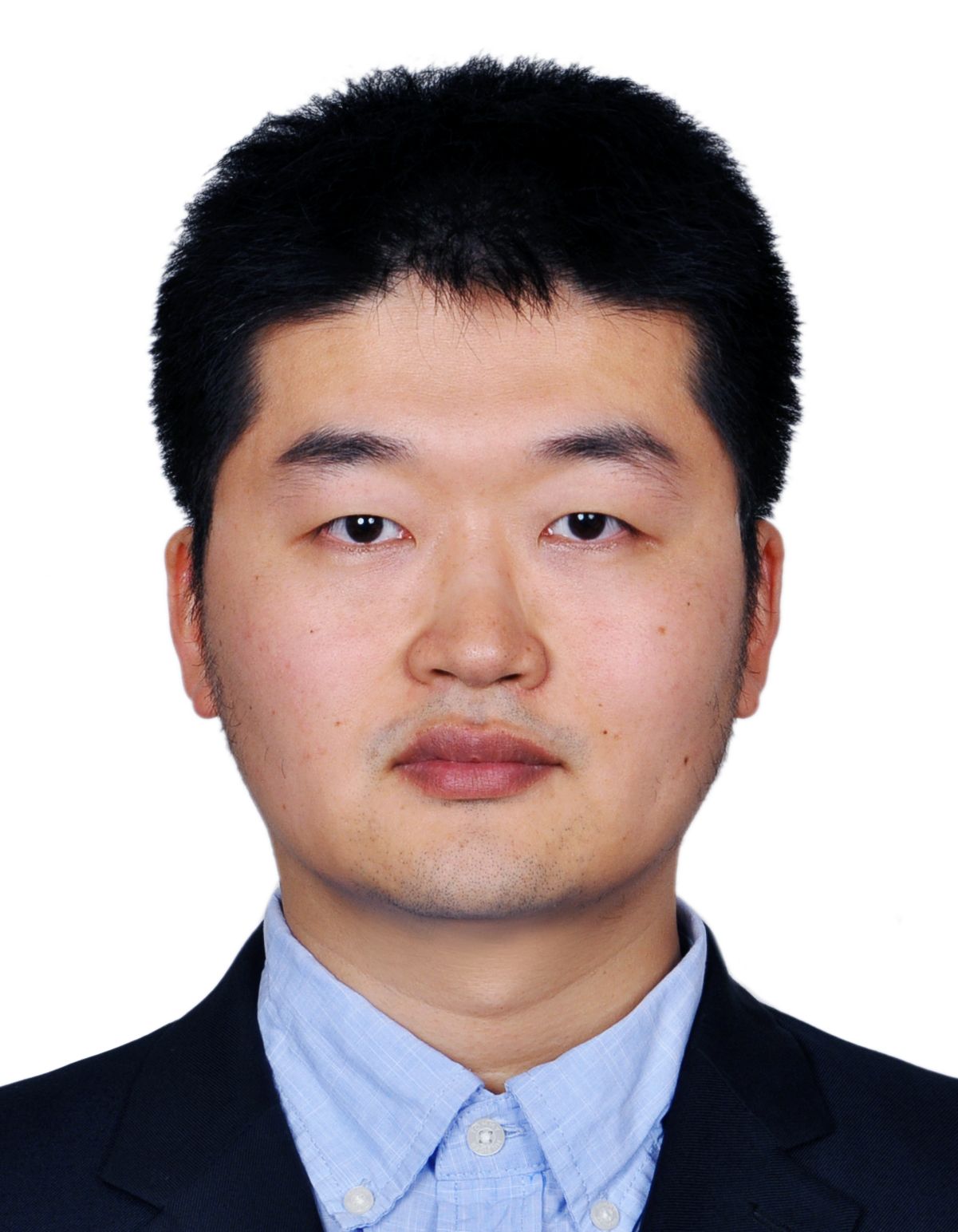}}]{Jia Li (Senior Member, IEEE) }
received the B.E. degree from Tsinghua University in 2005, and the Ph.D. degree from the Institute of Computing Technology, Chinese Academy of Sciences in 2011. He is currently a Full Professor with the School of Computer Science and Engineering, Beihang University, Beijing, China. Before he joined Beihang University in June 2014, he used to conduct research at Nanyang Technological University, Peking University, and Shanda Innovations. He is the author or coauthor of over 90 technical papers in refereed journals and conferences, such as IEEE TRANSACTIONS ON PATTERN ANALYSIS AND MACHINE INTELLIGENCE (TPAMI), IJCV, IEEE TRANSACTIONS ON IMAGE PROCESSING (TIP), CVPR, and ICCV. His research interests include computer vision and multimedia big data, especially the understanding and generation of visual contents. He is a Senior Member of ACM, CIE, and CCF. He has been supported by the Research Funds for Excellent Young Researchers from the National Natural Science Foundation of China since 2019. In 2017, he was selected into the Beijing Nova Program and ever received the Second-Grade Science Award of Chinese Institute of Electronics in 2018. He received the two Excellent Doctoral Thesis Award from the Chinese Academy of Sciences in 2012 and the Beijing Municipal Education Commission in 2012. He received the First-Grade Science-Technology Progress Award from the Ministry of Education, China, in 2010. 
\end{IEEEbiography}

\begin{IEEEbiography}[{\includegraphics[width=0.8in,height=1in,clip,keepaspectratio]{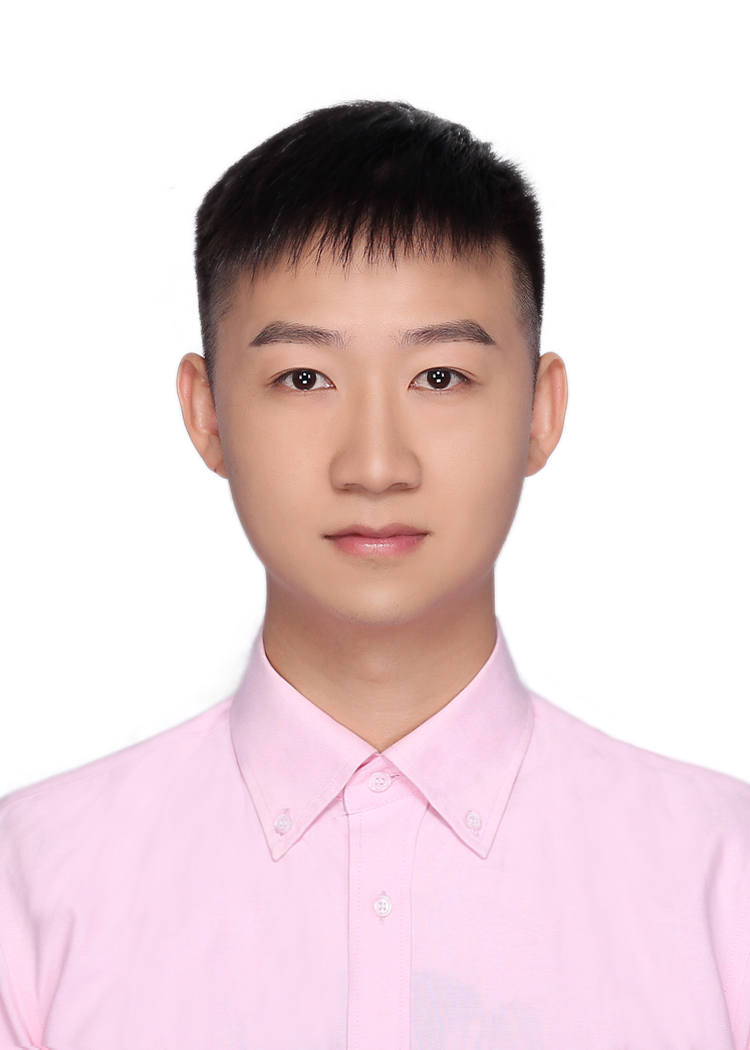}}]{Jinming Su}
received his master degree with the State Key Laboratory of Virtual Reality Technology and Systems, School of Computer Science and Engineering, Beihang University in Jan. 2020. He received the B.S. degree from
School of Computer Science and Engineering, Northeastern University, in Jul. 2017. His research interests include computer vision, visual saliency analysis and deep learning.
\end{IEEEbiography}

\begin{IEEEbiography}[{\includegraphics[width=0.8in,height=1in,clip,keepaspectratio]{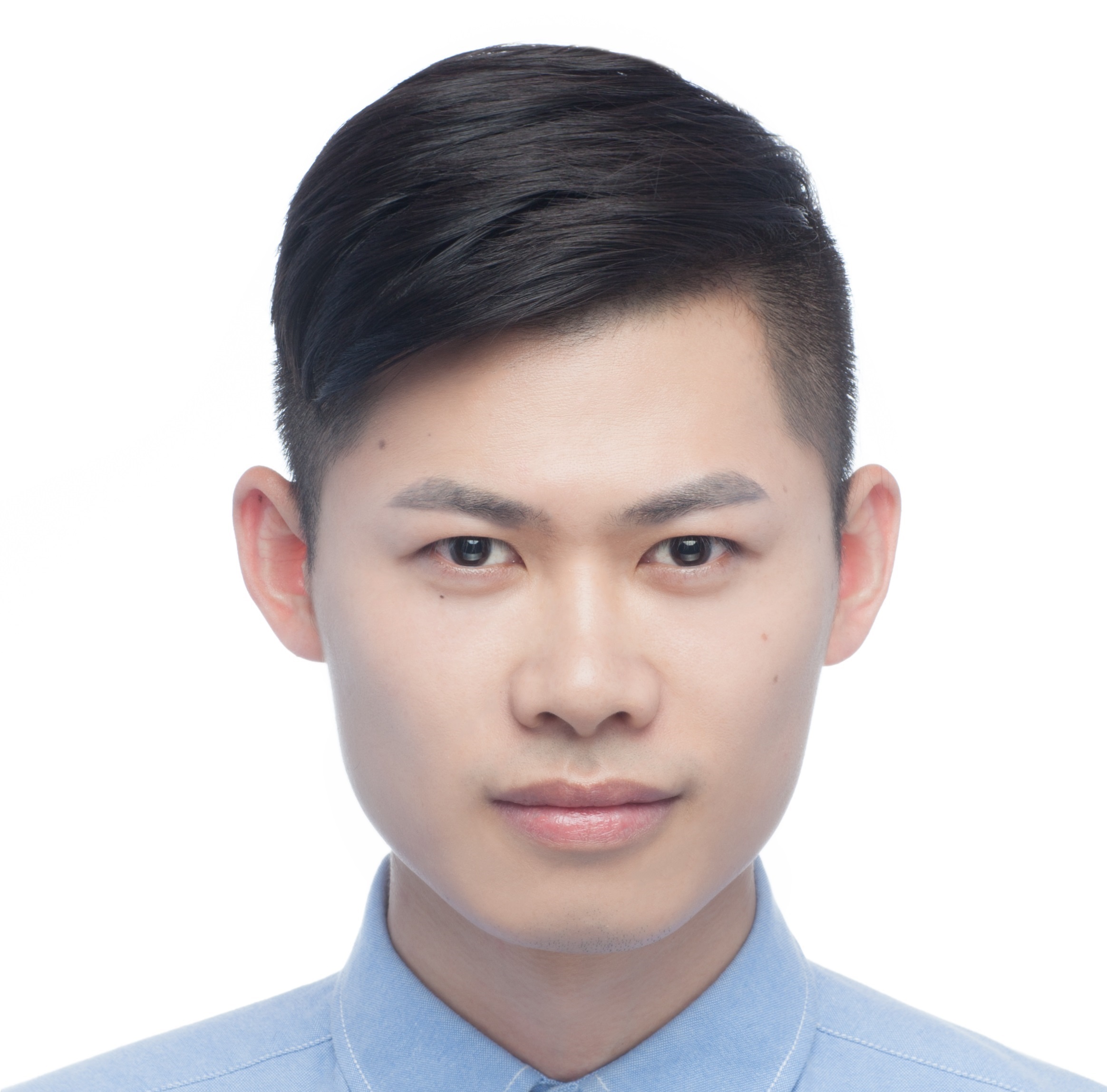}}]{Changqun Xia}
is currently an assistant Professor at Peng Cheng Laboratory, China. He received the Ph.D. degree from the State Key Laboratory of Virtual Reality Technology and Systems, School of Computer Science and Engineering, Beihang University, in Jul. 2019. His research interests include computer vision and image/video understanding.
\end{IEEEbiography}

\begin{IEEEbiography}[{\includegraphics[width=0.8in,height=1in,clip,keepaspectratio]{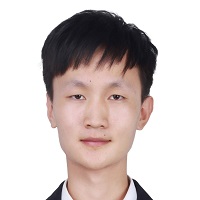}}]{Mingcan Ma}
is currently pursuing his master degree with the State Key Laboratory of Virtual Reality Technology and Systems, School of Computer Science and Engineering, Beihang University. His research interests include computer vision, image salient object detection and deep learning.
\end{IEEEbiography}

\begin{IEEEbiography}[{\includegraphics[width=0.8in,height=1in,clip,keepaspectratio]{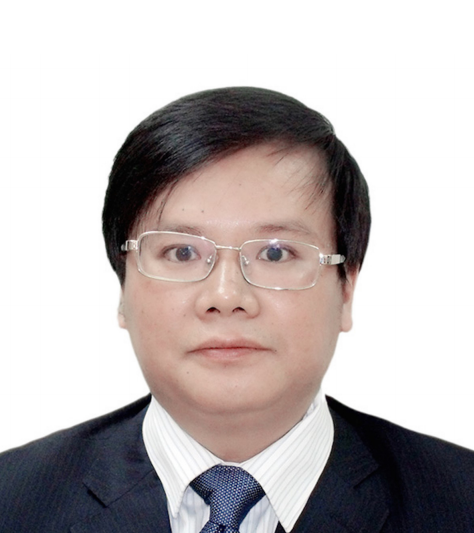}}]{Yonghong Tian (Senior Member, IEEE) }
received the Ph.D. degree from the Institute of Computing Technology, Chinese Academy of Sciences, Beijing, China, in 2005. He is currently a Full Professor with the National Engineering Laboratory for Video Technology, School of Electronics Engineering and Computer Science, Peking University, Beijing, China. He has authored or coauthored more than 160 technical articles in refereed journals and conferences, and has owned more than 57 Chinese and US patents. His research interests include machine learning, computer vision, and multimedia big data. Prof. Tian is a Senior Member of IEEE, CIE and CCF, and a Member of ACM. He is currently an Associate Editor of the IEEE TRANSACTIONS ON MULTIMEDIA, IEEE TRANSACTIONS ON CIRCUITS AND SYSTEMS FOR VIDEO TECHNOLOGY, IEEE MULTIMEDIA MAGAZINE, and IEEE ACCESS, and a co-Editor-in-Chief of the International Journal
of Multimedia Data Engineering and Management. He has served as the Technical Program Co-Chair of IEEE ICME 2015, IEEE BigMM 2015, IEEE ISM 2015 and IEEE MIPR 2018/2019, an Organizing Committee Member of more than ten conferences such as ACM Multimedia 2009, IEEE MMSP 2011, IEEE ISCAS 2013, and IEEE ISM 2016, and BigMMs 2018, and a PC Member or Area Chair of several conferences such as CVPR, ICCV, KDD, AAAI, ACM MM, ECCV, and ICME. He was the recipient of two national prizes and three ministerial prizes in China, and was the recipient of the 2015 EURASIP Best Paper Award for the EURASIP Journal on Image and Video Processing.
\end{IEEEbiography}

\end{document}